\documentclass[conference]{IEEEtran}
\IEEEoverridecommandlockouts
\usepackage{tcolorbox}
\usepackage{booktabs}
\usepackage[hidelinks,bookmarks=false]{hyperref}
\usepackage[font=scriptsize]{subcaption}
\usepackage{multirow}
\usepackage{xcolor}
\usepackage[percent]{overpic}
\def\BibTeX{{\rm B\kern-.05em{\sc i\kern-.025em b}\kern-.08em
    T\kern-.1667em\lower.7ex\hbox{E}\kern-.125emX}}
\newcommand{\ignore}[1]{}

\newcommand{\markus}[1]{\textcolor{blue}{\textbf{Markus:} #1}}

\begin{document}


\title{Predicting Good Configurations for\\GitHub and Stack Overflow Topic Models}

\author{\IEEEauthorblockN{Christoph Treude and Markus Wagner}
\IEEEauthorblockA{School of Computer Science\\University of Adelaide\\
\{christoph.treude$|$markus.wagner\}@adelaide.edu.au}}

\maketitle

\begin{abstract}
Software repositories contain large amounts of textual data, ranging from source code comments and issue descriptions to questions, answers, and comments on Stack Overflow. To make sense of this textual data, topic modelling is frequently used as a text-mining tool for the discovery of hidden semantic structures in text bodies. Latent Dirichlet allocation (LDA) is a commonly used topic model that aims to explain the structure of a corpus by grouping texts. LDA requires multiple parameters to work well, and there are only rough and sometimes conflicting guidelines available on how these parameters should be set. In this paper, we contribute (i) a broad study of parameters to arrive at good local optima for GitHub and Stack Overflow text corpora, (ii) an a-posteriori characterisation of text corpora related to eight programming languages, and (iii) an analysis of corpus feature importance via per-corpus LDA configuration. We find that (1) popular rules of thumb for topic modelling parameter configuration are not applicable to the corpora used in our experiments, (2) corpora sampled from GitHub and Stack Overflow have different characteristics and require different configurations to achieve good model fit, and (3) we can predict good configurations for unseen corpora reliably. These findings support researchers and practitioners in efficiently determining suitable configurations for topic modelling when analysing textual data contained in software repositories.
\end{abstract}


\vspace{-1mm}
\section{Introduction}

Enabled by technology, humans produce more text than ever before, and the productivity in many domains depends on how quickly and effectively this textual content can be consumed. In the software development domain, more than 8 million registered users have contributed more than 38 million posts on the question-and-answer forum Stack Overflow since its inception in 2008~\cite{SOTorrent}, and 67 million repositories have been created on the social developer site GitHub which was founded in the same year~\cite{CMU}. The productivity of developers depends to a large extent on how effectively they can make sense of this plethora of information.

The text processing community has invented many techniques to process large amounts of textual data, e.g., through topic modelling~\cite{Blei2003}. Topic modelling is a probabilistic technique to summarise large corpora of text documents by automatically discovering the semantic themes, or topics, hidden within the data. To make use of topic modelling, a number of parameters have to be set.

\begin{figure}[t]\centering\vspace{-2mm}%
\begin{minipage}[b]{\linewidth}%
\begin{overpic}[width=\textwidth]{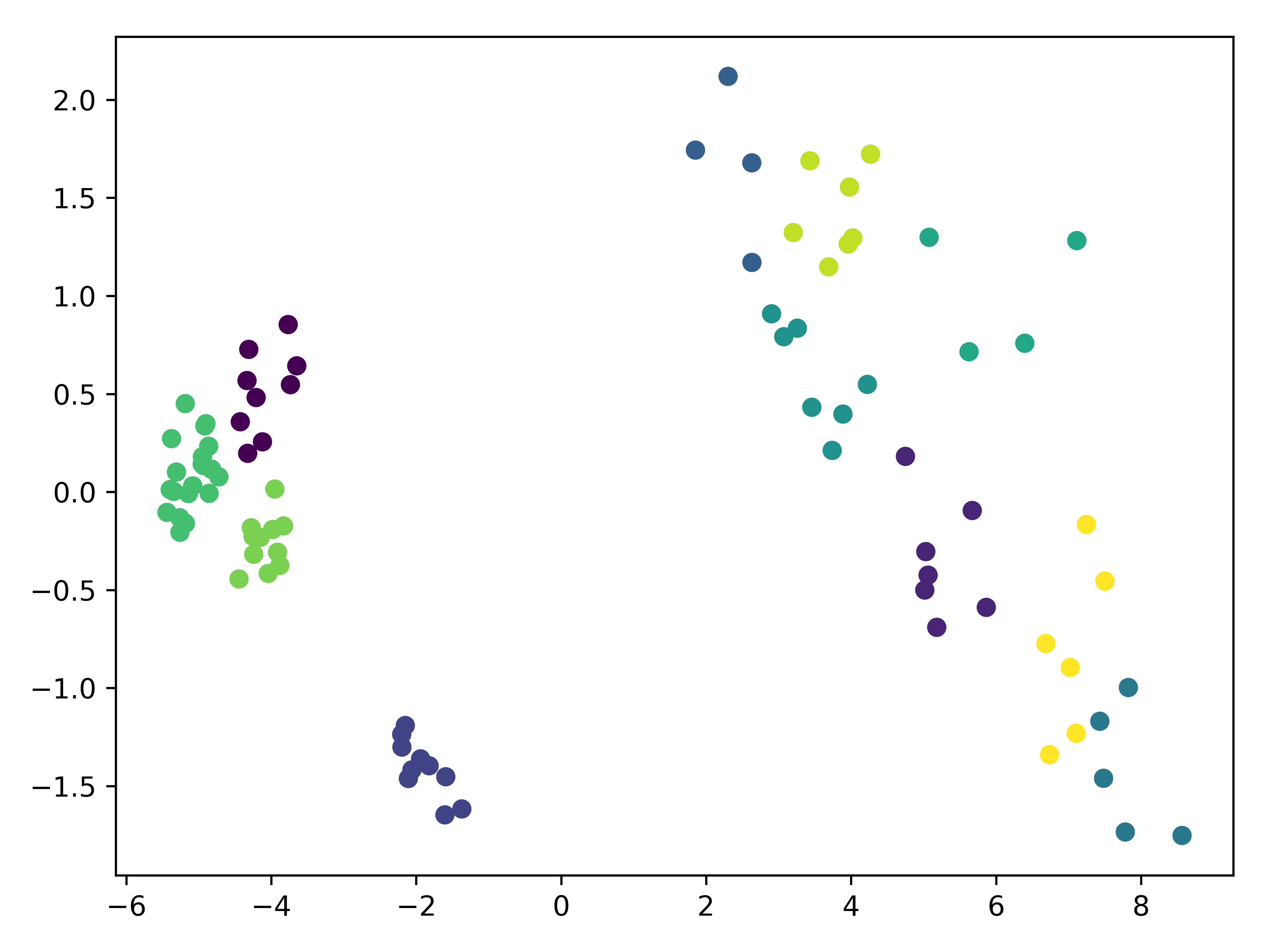}
 \put (15,28) {\scriptsize SO:Ruby}
 \put (28,8) {\scriptsize SO:C}
 \put (24,18) {\scriptsize SO:C++}
 \put (15,52) {\scriptsize SO:Java}
 \put (13,46) {\scriptsize SO:Python}
 \put (10,40) {\scriptsize SO:JavaScript}
 \put (10,34) {\scriptsize SO:CSS,HTML}
 \put (79,8) {\scriptsize GH:C}
 \put (74,16) {\scriptsize GH:C++}
 \put (70,32) {\scriptsize GH:Ruby}
 \put (61,36) {\scriptsize GH:Java}
 \put (60,46) {\scriptsize GH:Python}
 \put (49,65) {\scriptsize GH:CSS,HTML}
 \put (52,55) {\scriptsize GH:JavaScript}
\end{overpic}
\end{minipage}%
\vspace{-1mm}
\caption{Clustered corpora in 2d. The colour encodes the cluster assigned to each corpus. GH/SO refer to GitHub and Stack Overflow. The axes do not have any particular meaning in projections like these.}%
\label{fig:instanceclusters:GHandSO}\vspace{-2mm}%
\end{figure}
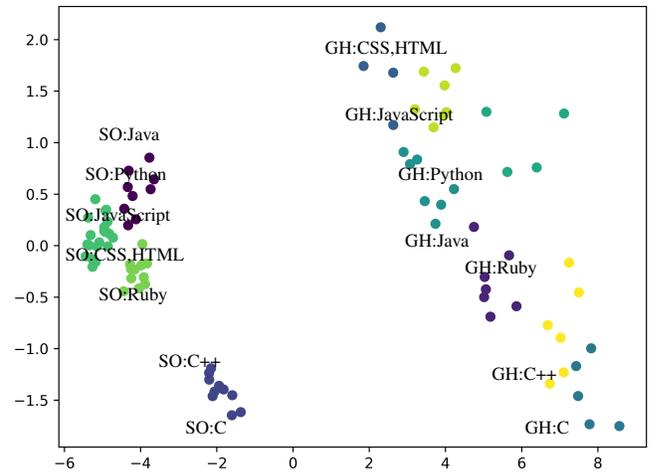

Agrawal et al.~\cite{Agrawal2016topicModelling} provide a recent overview of literature on topic modelling in software engineering. In the 24 articles they highlight, 23 of 24 mention instability in a commonly used technique to create topic models, i.e., with respect to the starting conditions and parameter choices. Despite this, all use default parameters, and only three of them perform tuning of some sort---all three use some form of a genetic algorithm. 

Even researchers who apply optimisation to their topic modelling efforts do not ``learn'' higher-level insights from their tuning, and there is very limited scientific evidence on the extent to which tuning depends on features of the corpora under analysis. For example, is the tuning that is needed for data from Stack Overflow different to the tuning needed for GitHub data? Does textual content related to some programming languages require different parameter settings compared to the textual content which discusses other programming languages? In this paper, we employ techniques from Data-Driven Software Engineering (DSE)~\cite{Nair2018DSE} and Data Mining Algorithms Using/Used-by Optimizers (DUO)~\cite{Agrawal2018duo} on 40 corpora sampled from GitHub and 40 corpora sampled from Stack Overflow to investigate the impact of per-corpus configuration on topic modelling. We ask two research questions:

\begin{description}
\item[RQ1] What are the optimal topic modelling configurations for textual corpora from GitHub and Stack Overflow?
\item[RQ2] Can we automatically select good configurations for unseen corpora based on their features alone?
\end{description}

We find that (1) popular rules of thumb for topic modelling parameter configuration are not applicable to textual corpora mined from software repositories, (2) corpora sampled from GitHub and Stack Overflow have different characteristics and require different configurations to achieve good model fit, and (3) we can predict good configurations for unseen corpora reliably based on corpus features. Figure~\ref{fig:instanceclusters:GHandSO} shows the corpora used in our experiments clustered in 2d based on their features after applying principal component analysis. The figure illustrates that textual corpora related to different programming languages and corpora taken from different sources (GitHub and Stack Overflow) can indeed be distinguished based on their features. Even across sources, the language-specific characteristics of the documents persist and corpora belonging to similar programming languages are close to each other. Moreover, the programming languages are in the vicinity of their spiritual ancestors and successors (e.g., C and C++).\footnote{See Section~\ref{sec:descrstats} for details.} We use this finding as a starting point for ad hoc per-corpus configuration of topic modelling of textual corpora mined from software repositories. Our predictions outperform the baseline by 4\% and are less than 1\% away from the virtual best solver.

These findings provide insight into the impact of corpus features on topic modelling in software engineering. They inform future work about efficient ways of determining suitable configurations for topic modelling, ultimately making it easier and more reliable for developers and researchers to understand the large amounts of textual data they are confronted with.

This article is structured as follows. First, we provide an introduction to topic modelling in Section~\ref{sec:tm}. 
Then, we describe in Section~\ref{sec:corpora} our data collection, and we provide a first statistical characterisation of the data.
In Section~\ref{sec:tuning}, we report on our tuning on individual corpora.
Section~\ref{sec:insights} provides insights gained from our per-corpus configuration and from the per-corpus parameter selection. Section~\ref{sec:threats} identifies threats which may affect the validity of our findings, before we discuss related work in Section~\ref{sec:related}. Finally, we conclude with a summary and by outlining future work.


\section{Topic Modelling}\label{sec:tm} 

Topic modelling is an information retrieval technique which automatically finds the overarching topics in a given text corpus, without the need for tags, training data, or predefined taxonomies~\cite{Barua2014}. Topic modelling makes use of word frequencies and co-occurrence of words in the documents in a corpus to build a model of related words~\cite{Blei2003}. Topic modelling has been applied to a wide range of artefacts in software engineering research, e.g., to understand the topics that mobile developers are talking about~\cite{Rosen2016}, to prioritise test cases~\cite{Thomas2014}, and to detect duplicate bug reports~\cite{Klein2014}.

The technique most commonly used to create topic models is Latent Dirichlet Allocation (LDA), a three-level hierarchical Bayesian model, in which each item of a collection is modelled as a finite mixture over an underlying set of topics~\cite{Blei2003}. A document's topic distribution is randomly sampled from a Dirichlet distribution with hyperparameter $\alpha$, and each topic's word distribution is randomly sampled from a Dirichlet distribution with hyperparameter $\beta$. $\alpha$ represents document-topic density---with a higher $\alpha$, documents contain more topics---while $\beta$ represents topic-word density---with a higher $\beta$, topics contain most of the words in the corpus~\cite{Luangaram2017}. In addition\ignore{ to $\alpha$ and $\beta$}, the number of topics---usually denoted as $k$---is another parameter needed to create a topic model using LDA. While many studies use the default settings for these parameters ($\alpha=1.0$, $\beta=0.01$, $k=100$; other sources suggest $\alpha=50/k$ and $\beta=0.1$~\cite{Griffiths2004}), in recent years, researchers have found that the defaults do not lead to the best model fit and have investigated the use of optimisation to determine good parameter values (e.g.,~\cite{Agrawal2016topicModelling}). To measure model fit, researchers have employed \textit{perplexity}, the geometric mean of the inverse marginal probability of each word in a held-out set of documents~\cite{Hoffman2010}, which we also use in this work. Low perplexity means the language model correctly guesses unseen words in test data.

In this work, we set out to investigate to what extent the optimal parameter settings for topic modelling depend on characteristics of the corpora being modelled. All our experiments were conducted with the LDA implementation Mallet, version 2.0.8.\footnote{\url{http://mallet.cs.umass.edu/download.php}, last accessed on 24 December 2018.}


\section{GitHub and Stack Overflow Corpora}\label{sec:corpora}

We now describe how we collected the documents used in our research. We define the features that we use to describe them, and we characterise them based on these features.

\subsection{Data Collection} 

\ignore{https://github.com/ctreude/topics/tree/master/StackOverflowCorpora10/split 
contains 500 files with corpora from Stack Overflow. 
These 500 belong to the 50 most popular programming languages on Stack Overflow, and there are 10 files with 1000 documents each for each programming language. 
https://github.com/ctreude/topics/blob/master/StackOverflowCorpora10/CorpusStatisticsStackOverflow.csv contains the corresponding corpus statistics.

https://github.com/ctreude/topics/tree/master/GitHubReadMeCorpora10/split 
https://github.com/ctreude/topics/blob/master/GitHubReadMeCorpora10/CorpusStatisticsGitHub.csv contains the corresponding corpus statistics.
contains 100 files with corpora from GitHub. 
}


To cover different sources and different content in our corpora, we sampled textual content related to eight programming languages from GitHub and Stack Overflow. We selected the set of languages most popular across both sources: C, C++, CSS, HTML, Java, JavaScript, Python, and Ruby.
As Stack Overflow has separate tags for HTML and HTML5 while GitHub does not distinguish between them, we considered both tags. Similarly, Stack Overflow distinguishes Ruby and Ruby-on-Rails, while GitHub does not.

For each programming language, we collected 5,000 documents which we stored as five corpora of 1,000 documents each to be able to generalise beyond a single corpus.
Our sampling and pre-processing methodology for both sources is described in the following. 

\textit{Stack Overflow sampling.} We downloaded the most recent 5,000 threads for each of the eight programming languages through the Stack Overflow API. Each thread forms one document (title + body + optional answers, separated by a single space). 

\textit{Stack Overflow pre-processing.} We removed line breaks (\verb+\n+ and \verb+\r+), code blocks (content surrounded by \verb+<pre><code>+), and all HTML tags from the documents. In addition, we replaced the HTML symbols \verb+&quot;+ \verb+&amp;+ \verb+&gt;+ and \verb+&lt;+ with their corresponding character, and we replaced strings indicating special characters (e.g., \verb+&#39;+) with double quotes. We also replaced sequences of whitespace with a single space.

\textit{GitHub sampling.} We randomly sampled \texttt{README.md} files of GitHub repositories that used at least one of the eight programming languages, using a script which repeatedly picks a random project ID between 0 and 120,000,000 (all GitHub repositories had an ID smaller than 120,000,000 at the time of our data collection). If the randomly chosen GitHub repository used at least one of the eight programming languages, we determined whether it contained a README file (cf.~\cite{README}) in the default location (\nolinkurl{https://github.com/<user>/<project>/blob/master/README.md}). If this README file contained at least 100 characters and no non-ASCII characters, we included its content as a document in our corpora.

\textit{GitHub pre-processing.} Similar to the Stack Overflow pre-processing, we removed line breaks, code blocks (content surrounded by at least 3 backticks), all HTML tags, single backticks, vertical and horizontal lines (often used to create tables), and comments (content surrounded by \verb+<!-- ... -->+). We also removed characters denoting sections headers (\# at the beginning of a line), characters that indicate formatting (*, \_), links (while keeping the link text), and badges (links preceded by an exclamation mark). In addition, we replaced the HTML symbols \verb+&quot;+ \verb+&amp;+ \verb+&gt;+ and \verb+&lt;+ with their corresponding character, and we replaced strings indicating special characters (e.g., \verb+&#39;+) with double quotes. We also replaced sequences of whitespace with a single space.


\subsection{Features of the Corpora}\label{sec:features}

\begin{table*}[t]
\centering
\caption{Features of Corpora. Features include the number of characters, words, and unique words as well as entropy, calculated separately for entire corpora and single documents.}
\begin{tabular}{cp{3.6cm}p{3.6cm}p{3.6cm}}
\toprule
& \multicolumn{3}{c}{scope} \\
\cmidrule{2-4}
& \multicolumn{1}{c}{corpus} & \multicolumn{1}{c}{document} & \multicolumn{1}{c}{document} \\
& & \multicolumn{1}{c}{(agg.~via median)} & \multicolumn{1}{c}{(agg.~via std dev)} \\
\midrule
\multirow{5}{*}{\rotatebox[origin=c]{90}{\centering \# characters}} & \textit{with stopwords:} \newline \texttt{\scriptsize{corpusCharacters}} \newline \textit{without stopwords:} \newline \texttt{\scriptsize{corpusCharacters-\newline NoStopwords}} & \textit{with stopwords:} \newline \texttt{\scriptsize{medianDocumentCharacters}} \newline \textit{without stopwords:} \newline \texttt{\scriptsize{medianDocumentCharacters-\newline NoStopwords}} & \textit{with stopwords:} \newline \texttt{\scriptsize{stdevDocumentCharacters}} \newline \textit{without stopwords:} \newline \texttt{\scriptsize{stdevDocumentCharacters-\newline NoStopwords}}\\
\midrule
\multirow{5}{*}{\rotatebox[origin=c]{90}{\centering \# words}} & \textit{with stopwords:} \newline \texttt{\scriptsize{corpusWords}} \newline \textit{without stopwords:} \newline \texttt{\scriptsize{corpusWords-\newline NoStopwords}} & \textit{with stopwords:} \newline \texttt{\scriptsize{medianDocumentWords}} \newline \textit{without stopwords:} \newline \texttt{\scriptsize{medianDocumentWords-\newline NoStopwords}} & \textit{with stopwords:} \newline \texttt{\scriptsize{stdevDocumentWords}} \newline \textit{without stopwords:} \newline \texttt{\scriptsize{stdevDocumentWords-\newline NoStopwords}}\\
\midrule
\multirow{5}{*}{\rotatebox[origin=c]{90}{\centering \parbox[c]{1.5cm}{\centering \# unique\newline words}}} & \textit{with stopwords:} \newline \texttt{\scriptsize{corpusUniqueWords}} \newline \textit{without stopwords:} \newline \texttt{\scriptsize{corpusUniqueWords-\newline NoStopwords}} & \textit{with stopwords:} \newline \texttt{\scriptsize{medianDocumentUniqueWords}} \newline \textit{without stopwords:} \newline \texttt{\scriptsize{medianDocumentUniqueWords-\newline NoStopwords}} & \textit{with stopwords:} \newline \texttt{\scriptsize{stdevDocumentUniqueWords}} \newline \textit{without stopwords:} \newline \texttt{\scriptsize{stdevDocumentUniqueWords-\newline NoStopwords}}\\
\midrule
\multirow{5}{*}{\rotatebox[origin=c]{90}{\centering entropy}} & \textit{with stopwords:} \newline \texttt{\scriptsize{corpusEntropy}} \newline \textit{without stopwords:} \newline \texttt{\scriptsize{corpusEntropy-\newline NoStopwords}} & \textit{with stopwords:} \newline \texttt{\scriptsize{medianDocumentEntropy}} \newline \textit{without stopwords:} \newline \texttt{\scriptsize{medianDocumentEntropy-\newline NoStopwords}} & \textit{with stopwords:} \newline \texttt{\scriptsize{stdevDocumentEntropy}} \newline \textit{without stopwords:} \newline \texttt{\scriptsize{stdevDocumentEntropy-\newline NoStopwords}}\\
\ignore{
\midrule
\rotatebox[origin=c]{90}{\centering source (2 features)} & \texttt{\scriptsize{source}} & -- & --\\
\midrule
\rotatebox[origin=c]{90}{\centering \parbox[c]{1.9cm}{\centering programming\newline language}} & \texttt{\scriptsize{language}} (8 features) & -- & --\\
}
\bottomrule
\end{tabular}
\label{tab:corpusfeatures}
\end{table*}

\ignore{
\begin{table*}[t]
\centering
\caption{Features of the corpora. Features include the number of characters, words, and unique words as well as entropy, calculated separately for entire corpora and single documents.}%
\begin{tabular}{llll}%
\toprule
& \multicolumn{3}{c}{scope} \\
\cmidrule{2-4}
& \multicolumn{1}{c}{corpus} & \multicolumn{1}{c}{document} & \multicolumn{1}{c}{document} \\
& & \multicolumn{1}{c}{(aggregated via median)} & \multicolumn{1}{c}{(aggregated via std dev)} \\
\midrule
\# characters & \texttt{\scriptsize{corpusCharacters}} & \texttt{\scriptsize{medianDocumentCharacters}} & \texttt{\scriptsize{stdevDocumentCharacters}}\\
\# words & \texttt{\scriptsize{corpusWords}} & \texttt{\scriptsize{medianDocumentWords}} & \texttt{\scriptsize{stdevDocumentWords}} \\
\# unique words & \texttt{\scriptsize{corpusUniqueWords}} & \texttt{\scriptsize{medianDocumentUniqueWords}} & \texttt{\scriptsize{stdevDocumentUniqueWords}}\\
entropy & \texttt{\scriptsize{corpusEntropy}} & \texttt{\scriptsize{medianDocumentEntropy}} & \texttt{\scriptsize{stdevDocumentEntropy}}\\
\bottomrule
\end{tabular}
\label{tab:corpusfeatures}
\end{table*}
}

We are not aware of any related work that performs per-corpus configuration of topic modelling and uses the features of a corpus to predict good parameter settings for a particular corpus. As mentioned before, Agrawal et al.~\cite{Agrawal2016topicModelling} found that only a small minority of the applications of topic modelling to software engineering data apply any kind of optimisation, and even the authors who apply optimisations do not ``learn'' higher-level insights from their experiments. While they all conclude that parameter tuning is important, it is unclear to what extent the tuning depends on corpus features. To enable such exploration, we calculated the 24 corpus features listed in Table~\ref{tab:corpusfeatures} (each feature is calculated twice, once with and once without taking into account stopwords\footnote{We used the ``Long Stopword List'' from \url{https://www.ranks.nl/stopwords}, last accessed on 24 December 2018.} to account for potential differences between feature values with and without stopwords, e.g., affecting the number of unique words).

We computed the number of characters in each entire corpus as well as the number of characters separately for each document in a corpus. To aggregate the number of characters per document to corpus level, we created separate features for their median and their standard deviation. This allowed us to capture typical document length in a corpus as well as diversity of the corpus in terms of document length. Similarly, we calculated the number of words and the number of unique words for each corpus and for each document. 

While these features capture the basic characteristics of a document and corpus in terms of length, they do not capture the nature of the corpus. To capture this, we relied on the concept of \textit{entropy}. As described by Koutrika et al.~\cite{Koutrika2015}, ``the basic intuition behind the entropy is that the higher a document's entropy is, the more topics the document covers hence the more general it is''. To calculate entropy, we used Shannon's definition~\cite{shannon1948mathematical}: 
\begin{equation}
-\sum_{i}p_i\log p_i
\end{equation}
where $p_i$ is the probability of word number $i$ appearing in the stream of words in a document. We calculated the entropy for each corpus and each document, considering the textual content with and without stopwords separately. Note that the runtime for calculating these values is at least $\Omega(n)$ since the frequency of each word has to be calculated separately.


\subsection{Descriptive Statistics}\label{sec:descrstats}

\begin{figure}\centering
\begin{minipage}[b]{\linewidth}%
\includegraphics[width=\linewidth,trim={0 0 10 143},clip]{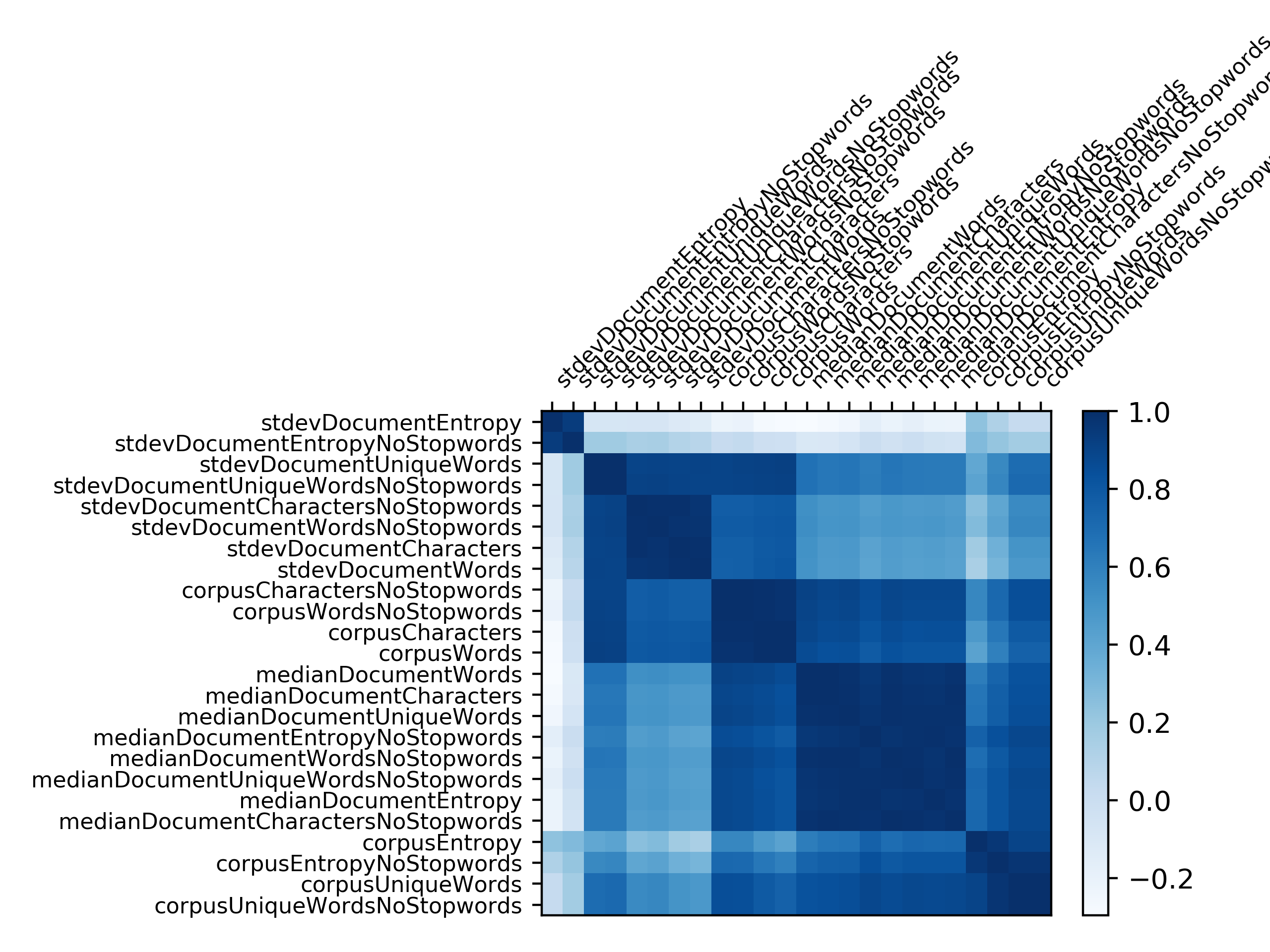}\vspace{-3mm}
\subcaption{GitHub corpora}\label{fig:featureCorrelations:GH}
\end{minipage}\\%
\begin{minipage}[b]{\linewidth}%
\includegraphics[width=\linewidth,trim={0 0 10 143},clip]
{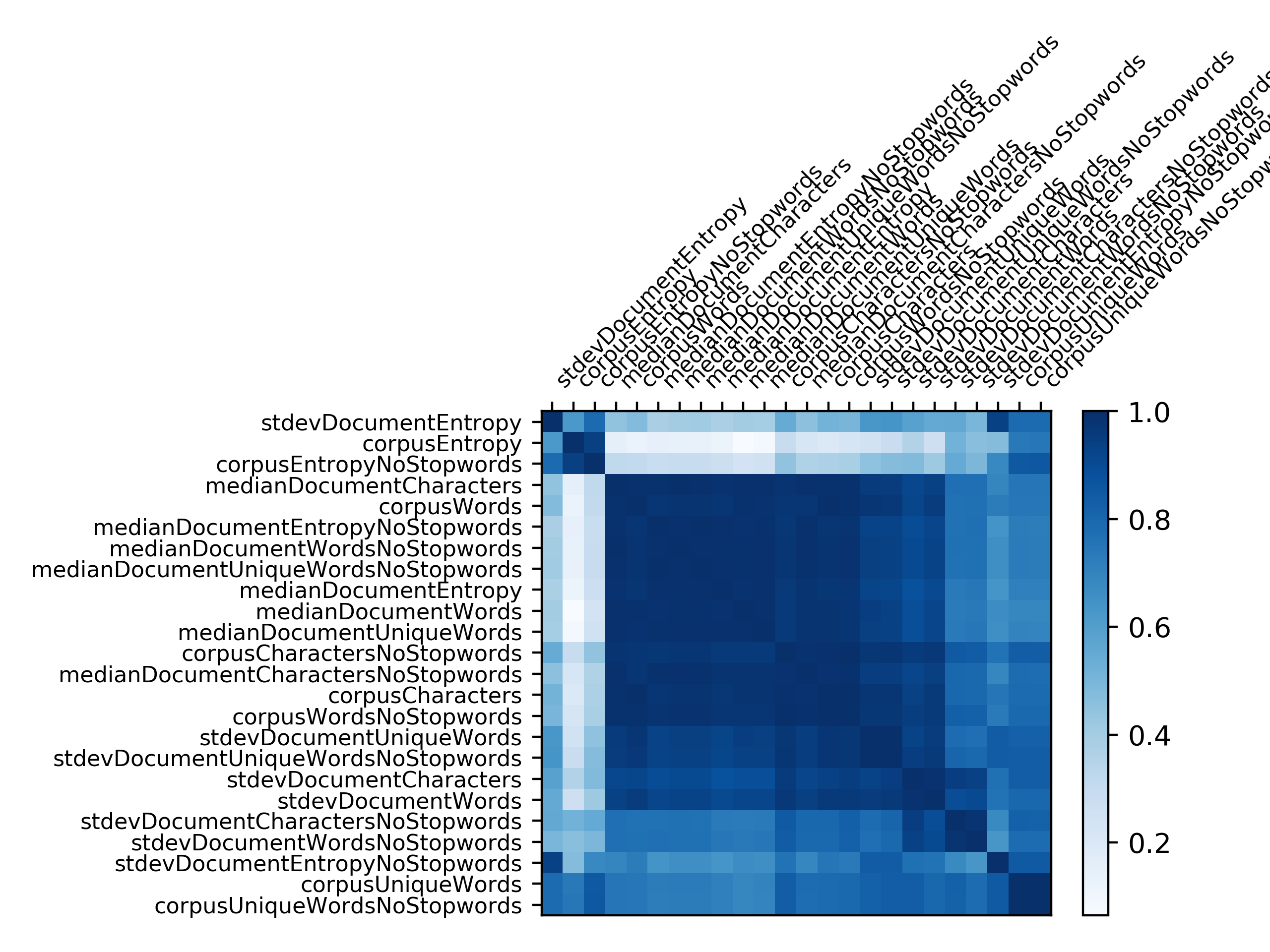}\vspace{-3mm}
\subcaption{Stack Overflow corpora}\label{fig:featureCorrelations:SO}
\end{minipage}\\
\begin{minipage}[b]{\linewidth}%
\includegraphics[width=\linewidth,trim={0 0 10 143},clip]
{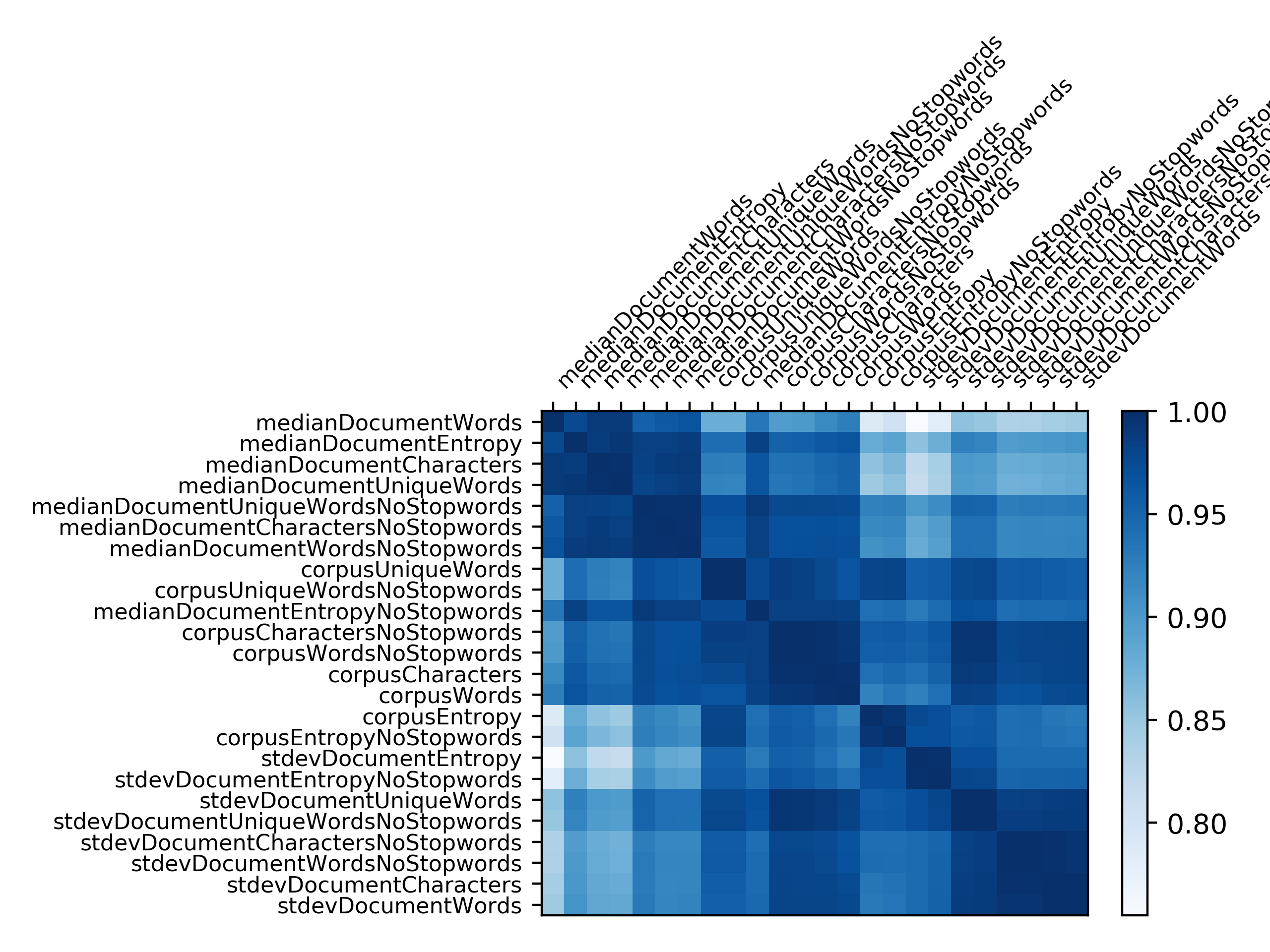}\vspace{-3mm}
\subcaption{Union}\label{fig:featureCorrelations:GHandSO}
\end{minipage}%
\caption{Correlations of features for each of the sources and for the union of GitHub and Stack Overflow corpora. Darker fields correspond to a larger correlation between the features. X-labels are omitted as they follow the order (optimised to co-locate correlated features) of the y-labels.}%
\label{fig:featureCorrelations}\vspace{-2mm}%
\end{figure}

While we have defined many corpus features, it is unclear how correlated these are, and whether the same relationships hold for GitHub README files and Stack Overflow discussions. Figure~\ref{fig:featureCorrelations} shows the correlations based on Pearson product-moment correlation coefficients between the 24 features and clustered with Wards hierarchical clustering approach.\footnote{Implementation provided by asapy~\cite{Bischl2016aslib}, \url{https://github.com/mlindauer/asapy}, last accessed on 24 December 2018.} As expected, the entropy-based features are correlated, as are those based on medians and standard deviations---this becomes particularly clear when we consider the relationships across all corpora (Figure~\ref{fig:featureCorrelations:GHandSO}). 

There are, however, differences between the two sources GitHub and Stack Overflow. For example, the stdevDocumentEntropy across the GitHub corpora is less correlated with the other features than among the Stack Overflow corpora. A reason for this could be that the README files from GitHub are different in structure from Stack Overflow threads. Also, the median-based feature values of the GitHub corpora are less correlated with the other features than in the Stack Overflow case. We conjecture this is because the README files vary more in length than in the Stack Overflow case, where thread lengths are more consistent.

Next, we will investigate differences between the programming languages. As we have 24 features and eight programming languages across two sources, we will limit ourselves to a few interesting cases here. 

\begin{figure}[t]
\hspace{18mm}\textsf{\scriptsize GitHub \hspace{13mm} Stack Overflow \hspace{14mm} Union}\\
\hspace*{5mm}
\rotatebox{90}{\hspace{5mm}\textsf{\scriptsize corpusWords}}
\begin{minipage}[b]{.3\linewidth}%
\includegraphics[width=\textwidth,height=21mm]{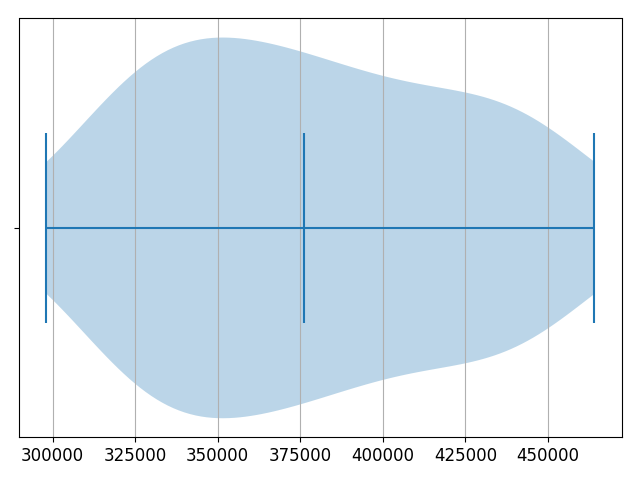}
\end{minipage}%
\begin{minipage}[b]{.3\linewidth}%
\includegraphics[width=\textwidth,height=21mm]
{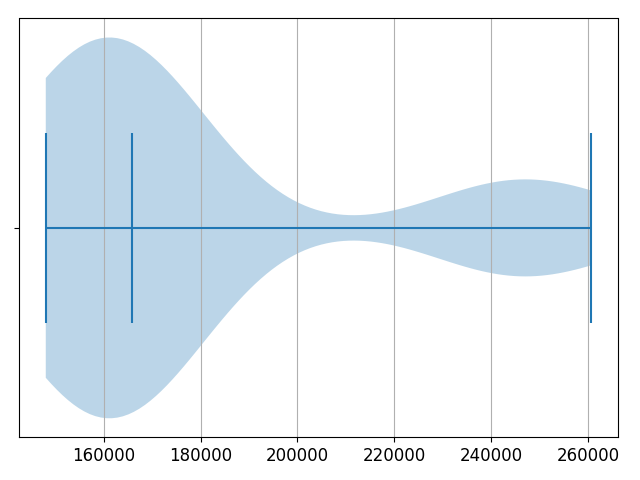}
\end{minipage}%
\begin{minipage}[b]{.3\linewidth}%
\includegraphics[width=\textwidth,height=21mm]{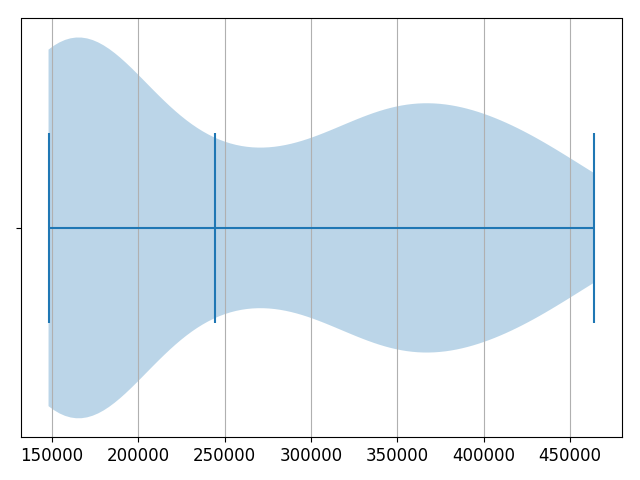}
\end{minipage}\\
\hspace*{1mm}
\rotatebox{90}{\hspace{5mm}\textsf{\scriptsize corpusWords-}}
\rotatebox{90}{\hspace{5mm}\textsf{\scriptsize NoStopwords}}
\begin{minipage}[b]{.3\linewidth}%
\includegraphics[width=\textwidth,height=21mm]{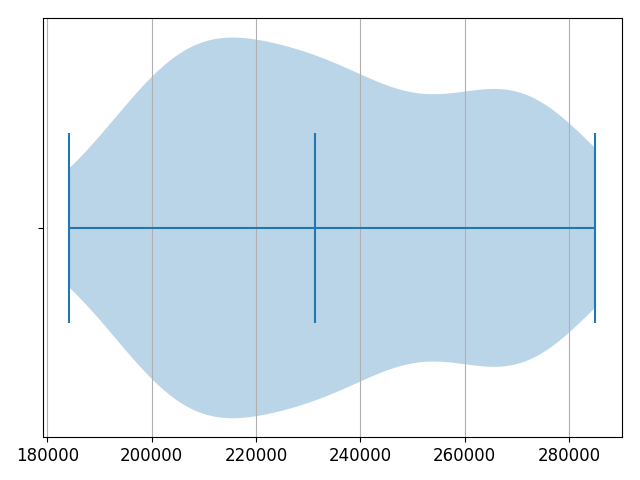}
\end{minipage}%
\begin{minipage}[b]{.3\linewidth}%
\includegraphics[width=\textwidth,height=21mm]
{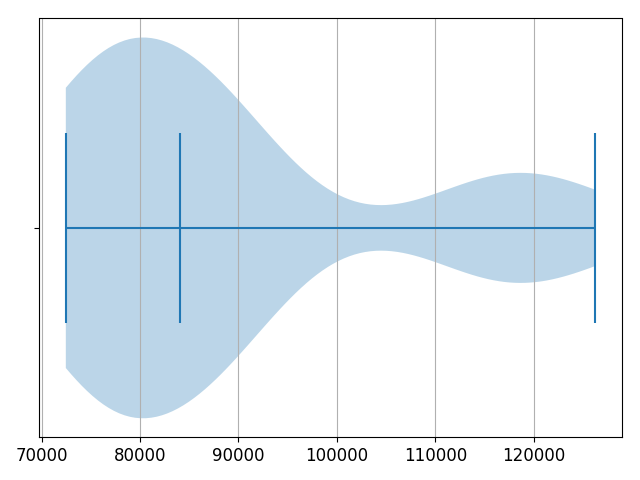}
\end{minipage}%
\begin{minipage}[b]{.3\linewidth}%
\includegraphics[width=\textwidth,height=21mm]{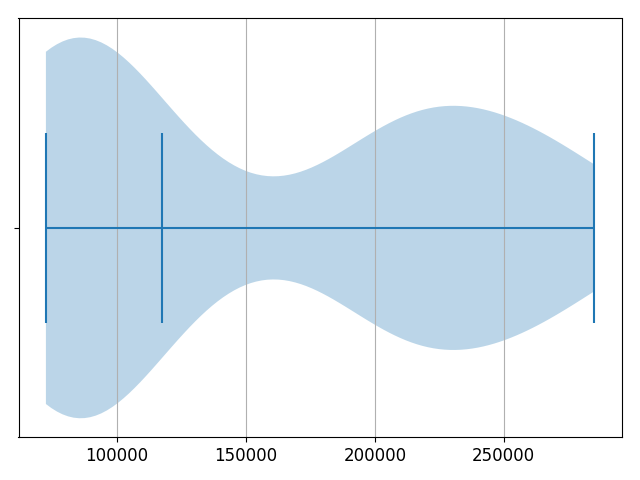}
\end{minipage}\\
\rotatebox{90}{\hspace{2mm}\textsf{\scriptsize medianDocument-}}
\rotatebox{90}{\hspace{3.5mm}\textsf{\scriptsize UniqueWords-}}
\rotatebox{90}{\hspace{4mm}\textsf{\scriptsize NoStopwords}}
\begin{minipage}[b]{.3\linewidth}%
\includegraphics[width=\textwidth,height=21mm]{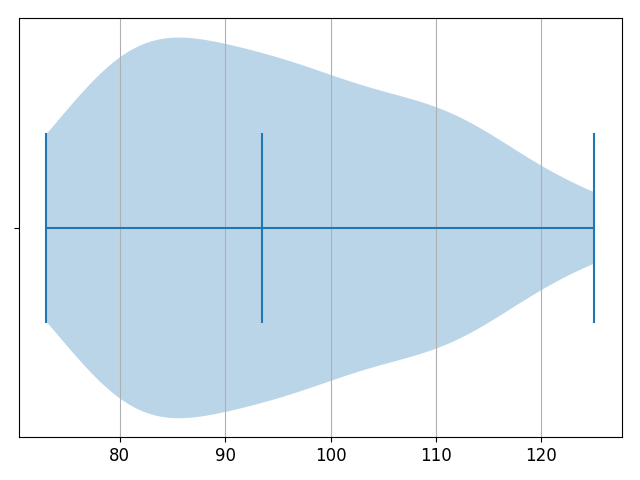}
\end{minipage}%
\begin{minipage}[b]{.3\linewidth}%
\includegraphics[width=\textwidth,height=21mm]
{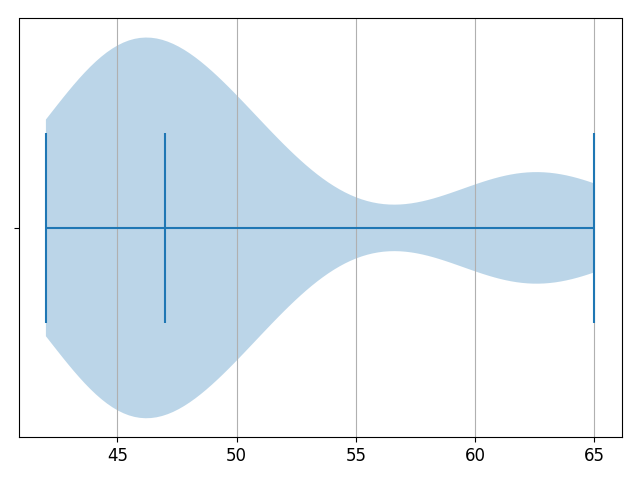}
\end{minipage}%
\begin{minipage}[b]{.3\linewidth}%
\includegraphics[width=\textwidth,height=21mm]{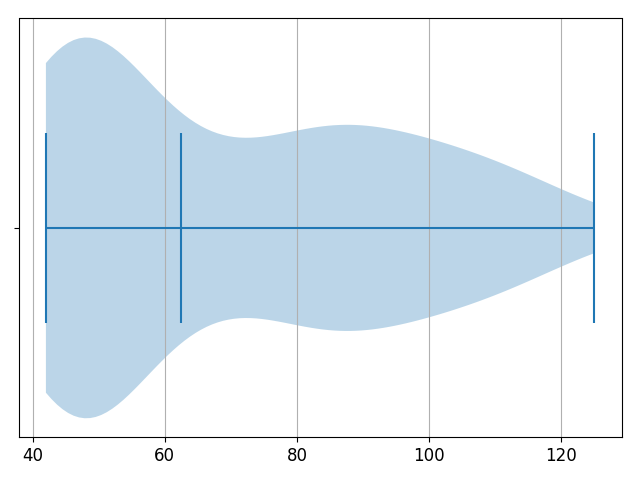}
\end{minipage}\\
%
\hspace*{2mm}
\rotatebox{90}{\hspace{4mm}\textsf{\scriptsize corpusEntropy-}}
\rotatebox{90}{\hspace{4.5mm}\textsf{\scriptsize NoStopwords}}
\begin{minipage}[b]{.3\linewidth}%
\includegraphics[width=\textwidth,height=23mm]{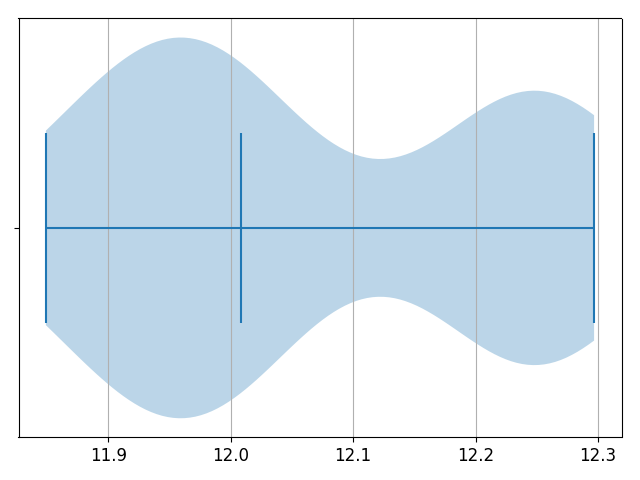}
\end{minipage}%
\begin{minipage}[b]{.3\linewidth}%
\includegraphics[width=\textwidth,height=23mm]
{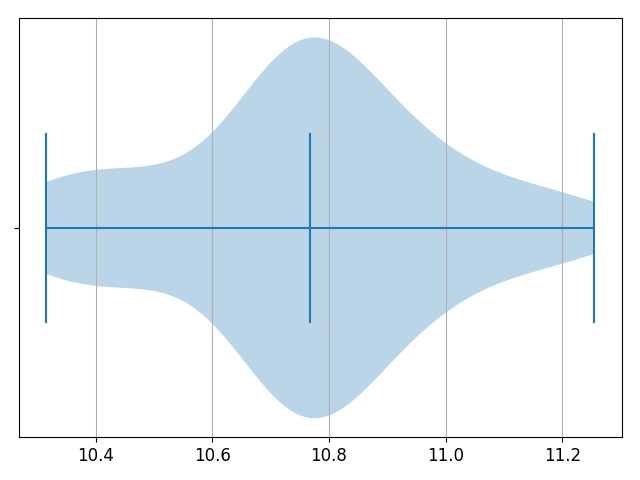}
\end{minipage}%
\begin{minipage}[b]{.3\linewidth}%
\includegraphics[width=\textwidth,height=23mm]{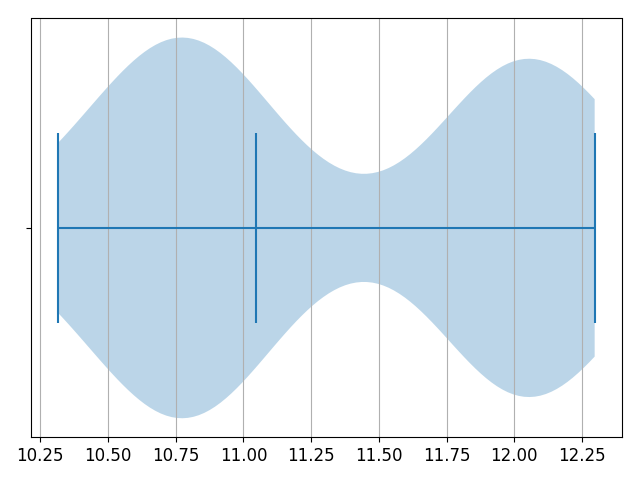}
\end{minipage}\\
\rotatebox{90}{\hspace{3mm}\textsf{\scriptsize medianDocument-}}
\rotatebox{90}{\hspace{8mm}\textsf{\scriptsize Entropy-}}
\rotatebox{90}{\hspace{5mm}\textsf{\scriptsize NoStopwords}}
\begin{minipage}[b]{.3\linewidth}%
\includegraphics[width=\textwidth,height=23mm]{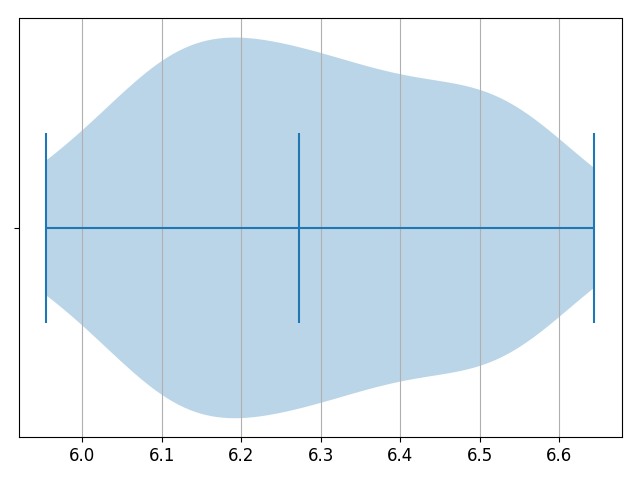}
\end{minipage}%
\begin{minipage}[b]{.3\linewidth}%
\includegraphics[width=\textwidth,height=23mm]
{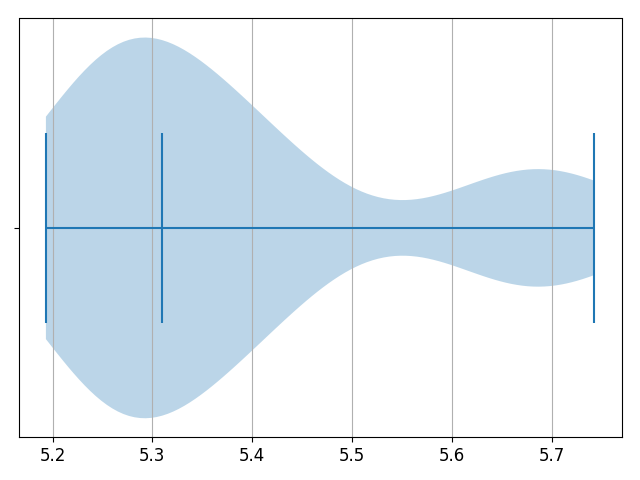}
\end{minipage}%
\begin{minipage}[b]{.3\linewidth}%
\includegraphics[width=\textwidth,height=23mm]{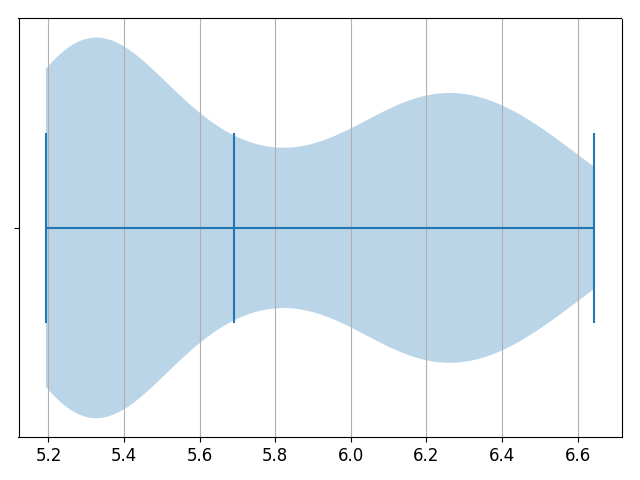}
\end{minipage}%
\caption{Characteristics of the corpora. Top: three features based on word counts; bottom: two features based on entropy. These violin plot are an alternative to box plots, and they indicate with thickness how common values are.}%
\label{fig:basicstats}\vspace{-2mm}%
\end{figure}

In Figure~\ref{fig:basicstats}, we start with a few easy-to-compute characteristics. For example, we see in the first row that GitHub documents are about twice as long as Stack Overflow discussions (see corpusWords). The distribution in the union shows this as well, with the left and the right humps (largely) coming from the two different sources. The trend remains the same if we remove stop words (see the second row). 
This already shows that we could tell the two sources apart with good accuracy by just considering either one of these easy-to-compute features. Despite this, the reliable classification of a single document does not appear to be as straightforward based on just the number of unique words that are not stop words: we can see in the third row that the two distributions effectively merged. 

Looking at entropy, which is significantly more time-consuming to compute, we can see the very same characteristics (see bottom two rows in Figure~\ref{fig:basicstats}). As seen before in Figure~\ref{fig:featureCorrelations}, entropy and word counts are correlated, but not as strongly with each other than some of the other measures.

Interestingly, GitHub documents contain fewer stop words (about 40\%) than Stack Overflow documents (almost 50\%). This seems to show the difference of the more technical descriptions present in the former in contrast to the sometimes more general discussion in the latter, which is reflected in the higher entropy of GitHub content compared to Stack Overflow content.

Next, we briefly investigate the heavy (right) tail of the Stack Overflow characteristics.
It turns out that this is caused by the C and C++ corpora. These are about 20-30\% longer than the next shorter ones on Ruby, Python, Java, with the shortest documents being on HTML, JavaScript and CSS. Roughly the same order holds for the entropy measures on the Stack Overflow data.

In the entropy characteristics of GitHub corpora, we note a bi-modal distribution. This time, Python joins C and C++ on the right-hand side, with all 15 corpora having a corpusEntropyNoStopwords value between 12.20 and 12.30. The closest is then a Java corpus with a value of 12.06. We speculate that software written in languages such as Python, Java, C, and C++ tends to be more complex than software written in HTML or CSS, which is reflected in the number of topics covered in the corresponding GitHub and Stack Overflow corpora measured in terms of entropy.

\begin{figure*}[t]\centering
\begin{minipage}[b]{.5\linewidth}%
\ignore{
\begin{overpic}[width=\textwidth,grid,tics=10]{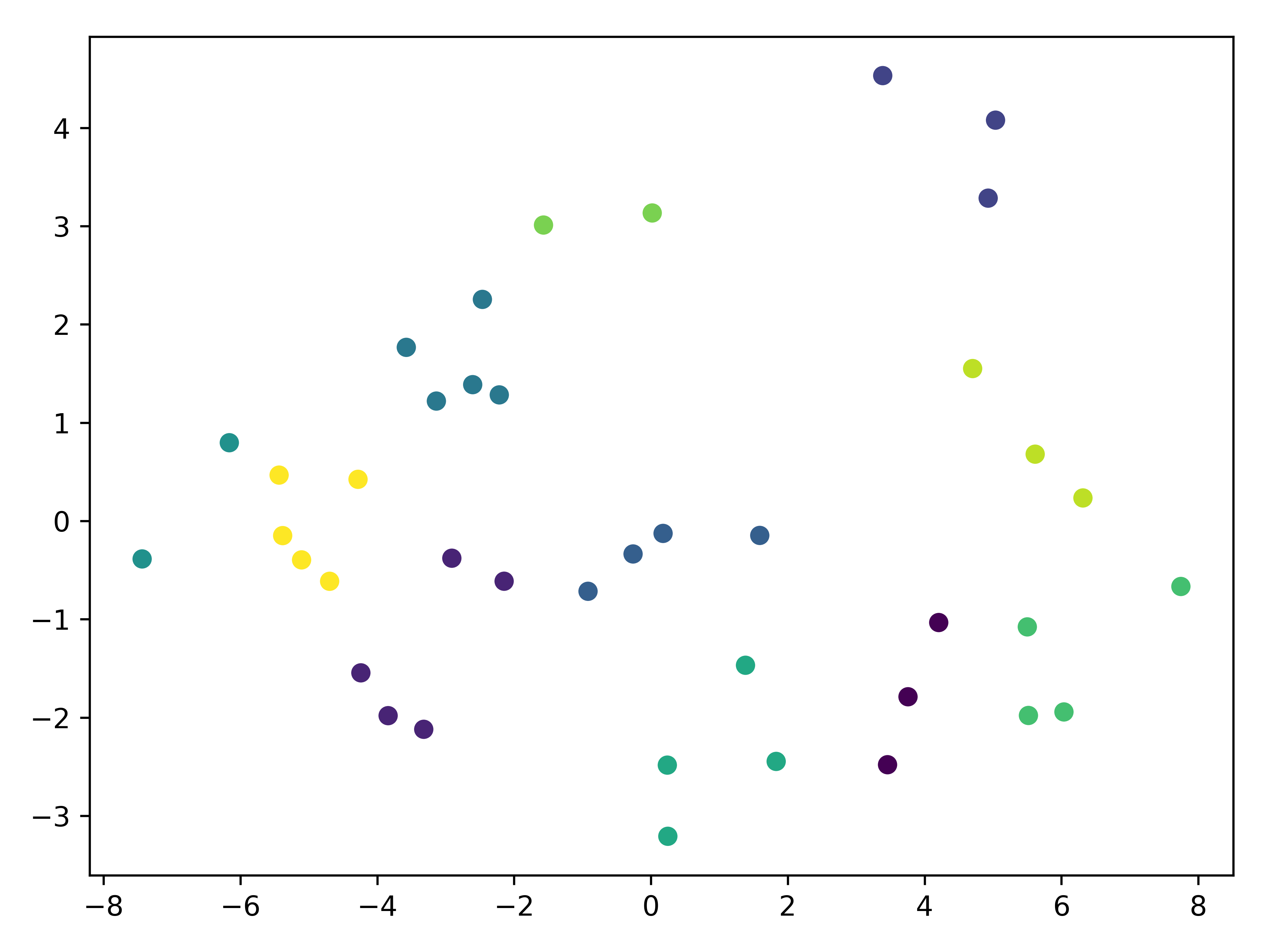}
 \put (20,85) {C}
\end{overpic}
}
\begin{overpic}[width=\textwidth]{180402GitHub_8_1to5_feature_clusters.png}
 \put (73,61) {\scriptsize Ruby}
 \put (80,30) {\scriptsize C}
 \put (72,22) {\scriptsize C++}
 \put (30,20) {\scriptsize Java}
 \put (22,43) {\scriptsize CSS}
 \put (40,50) {\scriptsize HTML}
 \put (47,16) {\scriptsize Python}
 \put (15,31) {\scriptsize JavaScript}
\end{overpic}\vspace{-1mm}%
\subcaption{GitHub corpora}\label{fig:instanceclusters:GH}
\end{minipage}%
\begin{minipage}[b]{.5\linewidth}%
\begin{overpic}[width=\textwidth]{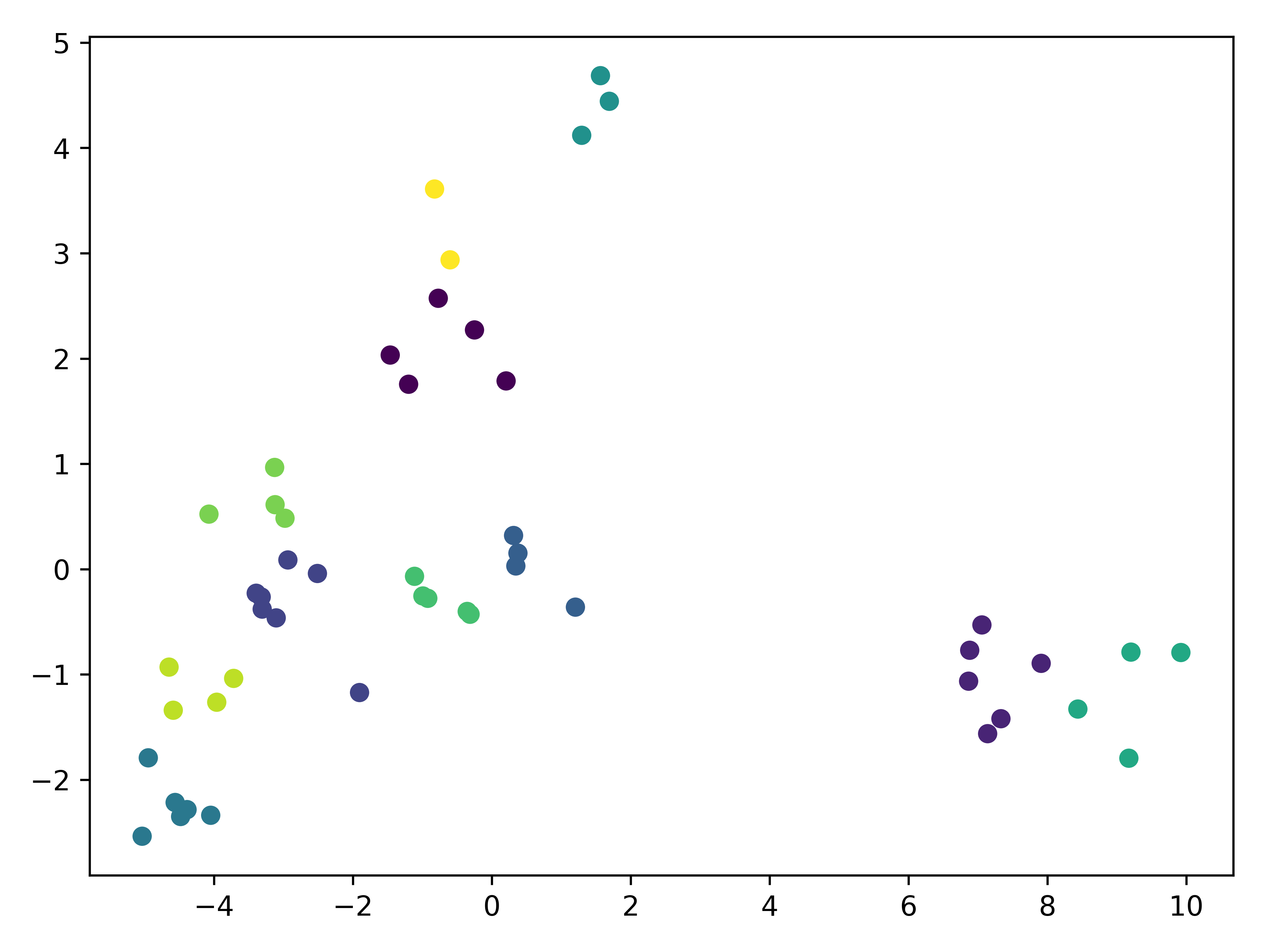}
 \put (42,30) {\scriptsize Ruby}
 \put (80,25) {\scriptsize C}
 \put (80,14) {\scriptsize C++}
 \put (35,60) {\scriptsize Java}
 \put (9,13) {\scriptsize CSS}
 \put (20,20) {\scriptsize HTML}
 \put (30,51) {\scriptsize Python}
 \put (20,40) {\scriptsize JavaScript}
\end{overpic}\vspace{-1mm}%
\subcaption{Stack Overflow corpora}\label{fig:instanceclusters:SO}
\end{minipage}%
\caption{Clustered corpora in 2d. The colour encodes the cluster assigned to each corpus. The dimensionality reductions for both sources were conducted independent of each other, thus resulting in arrangements different from the one in Figure~\ref{fig:instanceclusters:GHandSO}.}%
\label{fig:instanceclusters}
\end{figure*}

Lastly, we cluster the corpora in the feature space using a k-means approach. As pre-processing, we use standard scaling and a principal component analysis to two dimensions. To guess the number of clusters, we use the silhouette score on the range of 2 to 12 in the number of clusters. 
It turns out the individual languages per source can be told apart using this clustering almost perfectly (Figure~\ref{fig:instanceclusters}), and the two sources GitHub and Stack Overflow can be distinguished perfectly---we see this as a good starting point for ad hoc per-corpus configuration of topic modelling. Even across sources, the language-specific characteristics of the documents persist and similar languages are near each other (see Figure~\ref{fig:instanceclusters:GHandSO}). Moreover, the programming languages are in the vicinity of their spiritual ancestors and successors.


\section{Per-Corpus Offline Tuning}\label{sec:tuning}

\ignore{
\subsection{Problem characterisation}\label{sec:prelimexp}
\markus{there might be no room for this in this paper!}
prelim experiments: based on rule of thumb, vary topics, alpha, beta, iters for a few instances, similar to Christoph's plot from a while ago
}


Many optimisation methods can be used to tune LDA parameters. As mentioned before, three works identified in a recent literature review~\cite{Agrawal2016topicModelling} performed tuning, in particular, using genetic algorithms.

LDA is sensitive to the starting seed, 
and this noise can pose a challenge to many optimisation algorithms as the optimiser gets somewhat misleading feedback. 
Luckily, in recent years, many automated parameter optimisation methods have been developed and published as software packages. General purpose approaches include ParamILS~\cite{hutter2007paramils}, SMAC~\cite{hutter2011smac}, GGA~\cite{ansotegui2009gga}, and the iterated f-race procedure called irace~\cite{birattari2002irace}. 
The aim of these is to allow a wide range of parameters to be efficiently tested in a systematic way. For example, irace's procedure begins with a large set of possible parameter configurations, and tests these on a succession of examples. As soon as there is sufficiently strong statistical evidence that a particular parameter setting is sub-optimal, then it is removed from consideration (the particular statistical test used in the f-race is the Friedman test). In practice, a large number of parameter settings will typically be eliminated after just a few iterations, making this an efficient process.

To answer our first research question \textit{What are the optimal topic modelling configurations for textual corpora from GitHub and Stack Overflow?}, we use irace 2.3~\cite{birattari2002irace}.\footnote{The irace Package, \url{http://iridia.ulb.ac.be/irace}, last accessed on 24 December 2018.} 
We give irace a budget of 10,000 LDA runs, and we allow irace to conduct restarts if convergence is noticed. Each LDA run has a computation budget of 1,000 iterations, which is based on preliminary experiments to provide very good results almost independent of the CPU time budget. 
The LDA performance is measured in the perplexity (see Section~\ref{sec:tm}). 
In the final testing phase, the best configurations per corpus (as determined by irace) are run 101 times to achieve stable average performance values with a standard error of the mean of 10\%. 
In our following analyses, we consider the median of these 101 runs. 

Our parameter ranges are wider than what has been considered in the literature (e.g.,~\cite{Griffiths2004}), and are informed by our preliminary experiments: 
number of topics $k \in [3,1000]$, $\alpha \in [0.001, 200.0]$, $\beta \in [0.001, 200.0]$.
As an initial configuration that irace can consider we provide it with $k=100$, $\alpha=1.0$, and $\beta=0.01$, which are Mallet's default values.

This set of experiments is performed on a compute node with Intel(R) Xeon(R) E7-4870 CPUs with 1 TB RAM. 
Determining a well-performing configuration for each corpus takes 30-36 hours on a compute node with 80 cores with 80x-parallelisation. The total computation time required by the per-corpus optimisations is about 30 CPU years. 

\begin{figure*}[t]
\small{\begin{verbatim}
# Testing configurations:            (the first number is the configuration ID)
     topics  alpha   beta
288     628 10.474 55.835
847     550  4.958 68.056
1408    562  3.745 23.961
1558    556  3.884 21.293
1575    496 11.660 39.552
# Testing of elite configurations:   (medians of 101 independent runs)
     288   847  1408 1558  1575
   236.7 236.3 235.9  235 234.2
\end{verbatim}
\caption{irace results for the corpus \texttt{CGitHub-1}. The upper block lists the configurations returned by irace.}
\label{fig:iraceoutput}}
\end{figure*}

As an example, we show in Figure~\ref{fig:iraceoutput} the final output of irace when optimising the parameters for one of the five corpora related to C and taken from GitHub \texttt{CGitHub-1}. 
For comparison, the seeded default configuration achieves a median perplexity of 342.1.
The configuration evolved to one with a large number of topics, and a very large $\beta$ value. 
We observe that the perplexity values are very close to each to each other (at about 234 to 237, or 31\% below Mallet's default performance) even though the configurations vary. 

We show the results in Table~\ref{fig:tunedValues}. It turns out that the corpora from both sources and from the eight programming languages require different parameter settings in order to achieve good perplexity values---and thus good and useful ``topics''.
While the $\alpha$ values are at least (almost always) in the same order of magnitude as the seeded default configuration ($k=100$, $\alpha=1.0$, $\beta=0.01$), the $\beta$ values deviate significantly from it, as does the number of topics, confirming recent findings by Agrawal et al.~\cite{Agrawal2016topicModelling}.

For example, the numbers of topics addressed in the GitHub corpora is significantly higher (based on the tuned and averaged configurations for good perplexity values) than in the Stack Overflow corpora. This might be due to the nature of the README files of different software projects in contrast to potentially a more limited scope of discussions on Stack Overflow. Also, the Stack Overflow corpora appear to vary a bit more (standard deviation is 22\% of the mean) than the GitHub corpora (16\%).

When it comes to the different programming languages, we observe that the number of topics in Python / C / C++ is highest for the GitHub corpora, which appears to be highly correlated with the outstanding values of corpusEntropyNoStopwords of these corpora observed in Section~\ref{sec:descrstats}. Similarly, the corpora with lowest entropy (i.e., CSS / HTML / JavaScript) appear to require the smallest number of topics for good perplexity values.

Other interesting observations are that the $\beta$ values vary more among the Stack Overflow corpora. The $\alpha$ values are 
mostly comparable across the two sources.

\begin{table*}[t]
\centering\small
\caption{Results of tuning (number of topics $k$, $\alpha$, $\beta$) for eight programming languages from two sources, with the goal of minimising perplexity.}%
\vspace{-3mm}\label{fig:tunedValues}{\small
\begin{tabular}{ll|rrrrrr|rr}
\toprule
  &            & \multicolumn{2}{c}{$k$} & \multicolumn{2}{c}{$\alpha$}      & \multicolumn{2}{c}{$\beta$} &      \multicolumn{2}{|c}{perplexity} \\
source & langauge   & mean  & stdev & mean & stdev & mean       & stdev & mean  & stdev \\
\midrule
\multirow{9}{*}{GitHub}     & C          & 521.2 & 73.7  & 3.94 & 4.35  & 68.4       & 35.8  & 236.5 & 6.5   \\
       & C++        & 577.4 & 173.6 & 1.75 & 1.20  & 61.7       & 32.9  & 228.4 & 5.2   \\
       & CSS        & 455.4 & 34.1  & 1.52 & 0.82  & 36.7       & 16.0  & 236.7 & 7.8   \\
       & HTML       & 439.2 & 37.0  & 0.93 & 0.09  & 45.4       & 17.6  & 236.6 & 8.6   \\
       & Java       & 480.2 & 76.0  & 1.81 & 0.89  & 44.6       & 37.1  & 226.0 & 3.1   \\
       & JavaScript & 484.0 & 19.9  & 1.59 & 0.57  & 23.4       & 18.2  & 238.1 & 2.7   \\
       & Python     & 529.0 & 43.7  & 1.51 & 0.27  & 32.9       & 14.9  & 257.4 & 10.9  \\
       & Ruby       & 505.4 & 28.0  & 2.41 & 1.49  & 89.1       & 37.0  & 213.9 & 6.0   \\ \cmidrule{2-10}
       & \textit{all}        & 499.0 & 81.0  & 1.93 & 1.80  & 50.3       & 32.4  & 234.2 & 13.3  \\
\midrule
\multirow{9}{*}{Stack Overflow}     & C          & 377.0 & 34.3  & 0.95 & 0.35  & 51.8       & 55.1  & 202.9 & 4.5   \\
       & C++        & 337.6 & 29.6  & 3.33 & 3.30  & 97.4       & 61.8  & 199.3 & 3.0   \\
       & CSS        & 196.2 & 24.2  & 1.01 & 0.96  & 18.1       & 15.3  & 184.1 & 2.7   \\
       & HTML       & 244.4 & 18.1  & 2.45 & 2.33  & 76.4       & 69.5  & 196.7 & 5.9   \\
       & Java       & 349.8 & 49.1  & 0.85 & 0.46  & 10.0       & 8.2   & 223.9 & 2.5   \\
       & JavaScript & 252.8 & 34.5  & 4.24 & 3.66  & 50.9       & 44.0  & 213.6 & 2.0   \\
       & Python     & 295.8 & 47.3  & 1.10 & 0.18  & 67.6       & 78.6  & 229.0 & 4.0   \\
       & Ruby       & 269.3 & 33.1  & 2.11 & 2.72  & 64.0       & 52.4  & 215.9 & 7.3   \\ \cmidrule{2-10}
       & \textit{all}        & 283.7 & 61.9  & 2.06 & 2.37  & 57.6       & 57.4  & 207.8 & 14.2  \\
\midrule
\multicolumn{2}{c}{\textit{all}}            & 379.4 & 128.7 & 2.00 & 2.12  & 54.4       & 47.8  & 219.5 & 19.1 \\
\bottomrule
\end{tabular}}\vspace{-1mm}
\end{table*}

\begin{tcolorbox}
\textbf{Summary}: Popular rules of thumb for topic modelling parameter configuration are not applicable to textual corpora from GitHub and Stack Overflow. These corpora have different characteristics and require different configurations to achieve good model fit.
\end{tcolorbox}


\section{Per-Corpus Configuration}\label{sec:insights}

An alternative to the tuning of algorithms is that of selecting an algorithm from a portfolio or determining an algorithm configuration, when an instance is given. This typically involves the training of machine learning models on performance data of algorithms in combination with instances given as feature data. In software engineering, this has been recently used as an approach for the Software Project Scheduling Problem~\cite{Shen2018,wu2016}. The field of per-instance configuration has received much attention recently, and we refer the interested reader to a recent updated survey article~\cite{kotthoff2016survey}. 
The idea of algorithm selection is that given an instance, an algorithm selector selects a well-performing algorithm from a (often small) set of algorithms, the so-called portfolio.

To answer our second research question \textit{Can we automatically select good configurations for unseen corpora based on their features alone?}, we study whether we can apply algorithm selection to LDA configuration to improve its performance further than with parameter tuning only. 
We take from each language and each source the tuned configuration of each first corpus (sorted alphabetically), and we consider our default configuration, resulting in a total of 17 configurations named gh.C, ... so.C, ... and default. As common in the area of algorithm portfolios, we treat these different configurations as different algorithms and try to predict which configuration should be used for a new given instance---``new'' are now all corpora from both sources. Effectively, this will let us test which tuned corpus-configuration performs well on others. 
A similar approach was used by Wagner et al. to investigate the importance of instance features in the context of per-instance configuration of solvers for the minimum vertex cover problem~\cite{Wagner2017mvctuning}, for the traveling salesperson problem~\cite{Nallaperuma2015frontiers}, and for the traveling thief problem~\cite{Wagner2017ttpstudy}.

\begin{figure}[t]\centering\vspace{-2mm}%
\includegraphics[width=\linewidth,trim={0 0 0 0},clip]{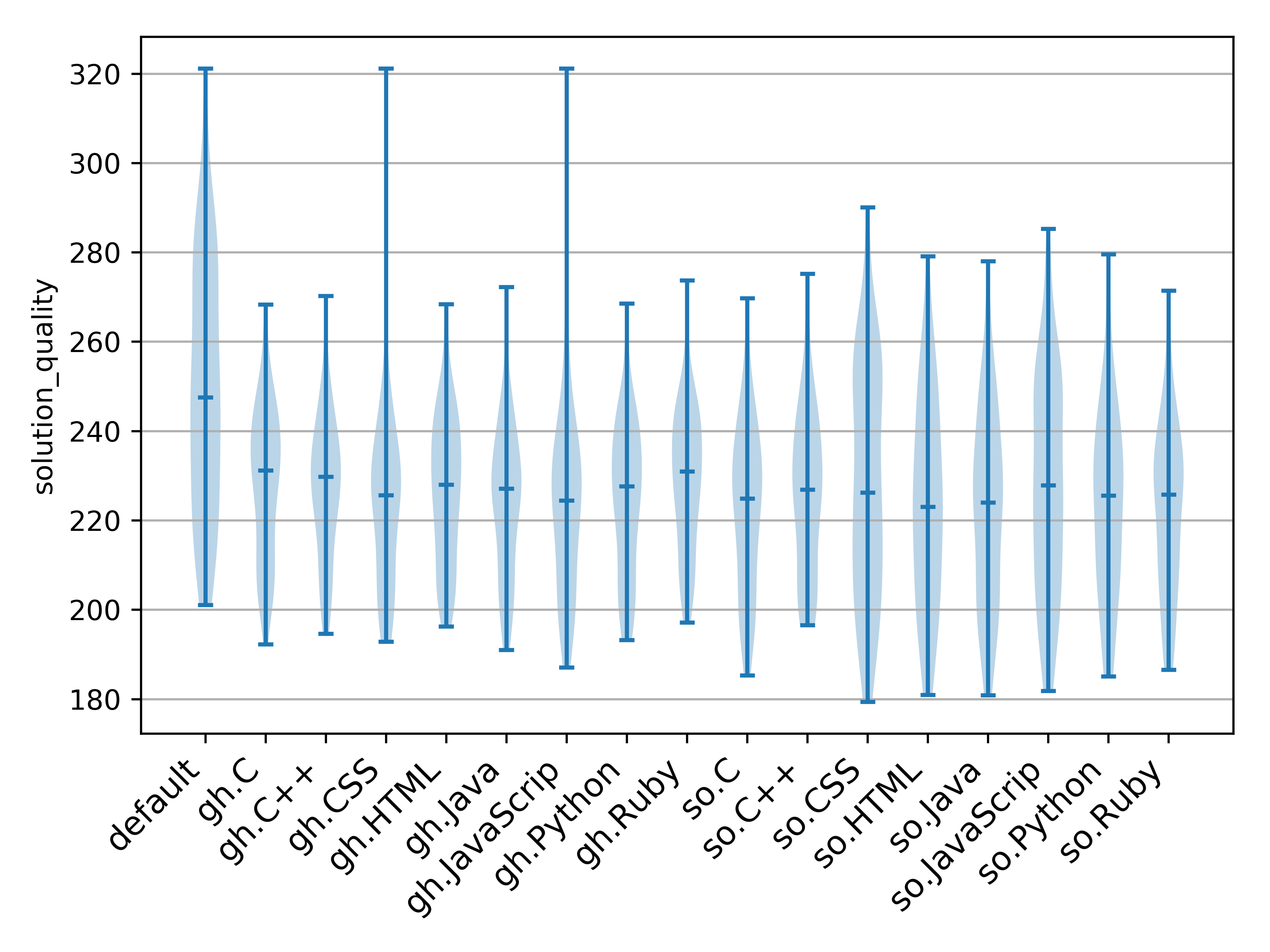}\vspace{-1mm}
\caption{Results of 17 given configurations across all corpora.}\label{fig:potfolioindividual}%
\label{fig:portfolio}\vspace{-2mm}%
\end{figure}%

As algorithm selection is often implemented using machine learning~\cite{Smithmiles2008as,Kerschke2019algorithmselection}, we need two preparation steps: (i) instance features that characterise instances numerically, (ii) performance data of each algorithm on each instance. We have already characterised our corpora in Section~\ref{sec:features}, so we only need to run each of the 17 configurations on all corpora.

\begin{figure}[t]\centering%
\begin{minipage}[b]{\linewidth}%
\begin{overpic}[width=\textwidth]{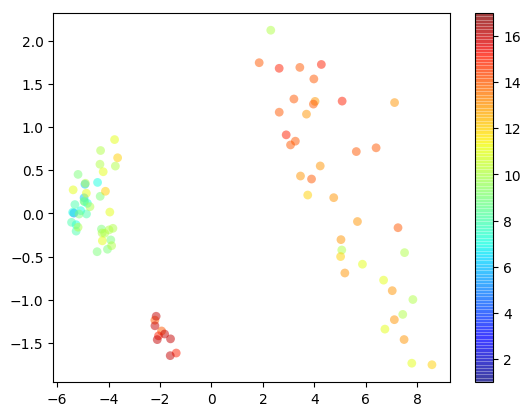}
 \put (15,28) {\scriptsize SO:Ruby}
 \put (28,8) {\scriptsize SO:C}
 \put (26,16) {\scriptsize SO:C++}
 \put (15,52) {\scriptsize SO:Java}
 \put (13,46) {\scriptsize SO:Python}
 \put (10,40) {\scriptsize SO:JavaScript}
 \put (10,34) {\scriptsize SO:CSS,HTML}
 \put (68,8) {\scriptsize GH:C}
 \put (63,16) {\scriptsize GH:C++}
 \put (62,25) {\scriptsize GH:Ruby}
 \put (61,38) {\scriptsize GH:Java}
 \put (58,45) {\scriptsize GH:Python}
 \put (49,65) {\scriptsize GH:CSS,HTML}
 \put (52,55) {\scriptsize GH:JavaScript}
\end{overpic}
\end{minipage}%
\caption{Results of per-corpus configuration.
Hardness: projected into 2d feature space (see Section~\ref{sec:descrstats}), the colour encodes the number of configurations that perform within 5\% of the best performance. The arrangement of instances is identical to that in Figure~\ref{fig:instanceclusters:GHandSO}.}%
\label{fig:potfoliocorr}
\end{figure}
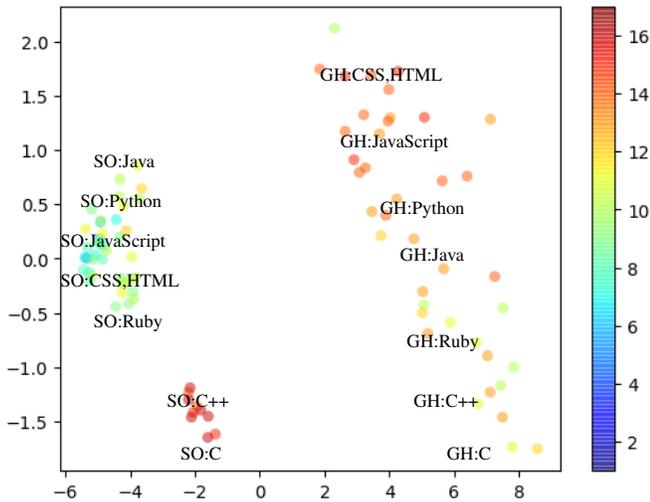%

Figure~\ref{fig:portfolio} provides an overview of the performance of the 17 configurations when run across all corpora.\footnote{gh.CSS and gh.JavaScript crashed on two corpora: we assigned as a result the maximum value observed (321.2).} As we can see, a per-corpus configuration is necessary to achieve the lowest perplexity values in topic modelling (Figure~\ref{fig:potfolioindividual}). Many configuration corpora can be optimised (within 5\%) with a large number of configurations (Figure~\ref{fig:potfoliocorr}, red), however, a particular cluster of Stack Overflow corpora requires specialised configurations.  

The average perplexity of the 17 configurations is 227.3. 
The single best configuration across all data is so.Java (tuned on one of the five Stack Overflow Java corpora) with an average perplexity value of 222.9; the default configuration achieves an average of 250.3 (+12\%). 

Based on all the data we have, we can simulate the so-called virtual best solver, which would pick for each corpus the best out of the 17 configurations. This virtual best solver has an average perplexity of 217.9, which is 2\% better than so.Java and 14\% better than the default configuration.



Lastly, let us look into the actual configuration selection. Using the approach of SATZilla'11~\cite{xu-rcra11a} as implemented in AutoFolio~\cite{Lindauer2015autofolio}, we train a cost-sensitive random forest for each pair of configurations, which then predicts for each pair of configurations the one that will perform better. The overall model then proposes the best-performing configuration. In our case, we use this approach to pick one of the 17 configurations given an instance that is described by its features. The trained model's predictions achieve an average perplexity of 219.6: this is a 4\% improvement over the average of the 17 tuned configurations, and it is less than 1\% away from the virtual best solver.

\begin{figure}[t]\centering
\includegraphics[width=\linewidth,trim={0 0 0 0},clip]{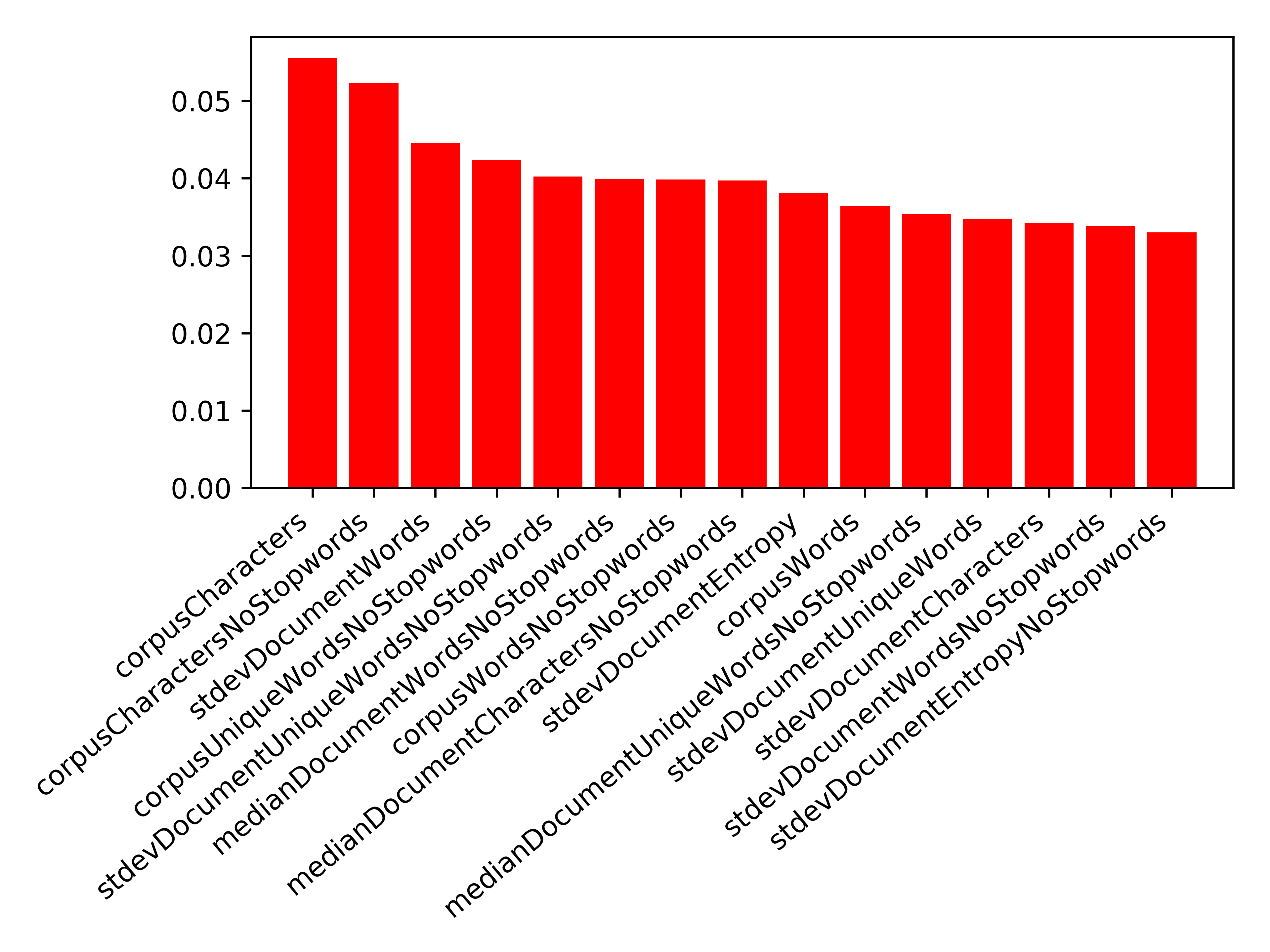}
\caption{Gini importance, features from Table~\ref{tab:corpusfeatures}.}
\label{fig:featureimportance}
\end{figure}

We are interested in the importance of features in the model---not only to learn about the domain, but also as the calculation of instance features forms an important step in the application of algorithm portfolios. The measure we use is the Gini importance~\cite{breimann-mlj01a} across all cost-sensitive random forests models, that can predict for a pair of solvers which one will perform better~\cite{xu-rcra11a}. 
Figure~\ref{fig:featureimportance} reveals that there is not a single feature, but a large set of features which together describe a corpus. 
It is therefore hardly possible to manually come up with good ``rules of thumb''  
to choose the appropriate configuration depending on the corpus features---even though many of the features are correlated (see Section~\ref{sec:descrstats}). 

Interestingly, the expensive-to-compute entropy-based features are of little importance in the random forests (1x 9th, 1x 15th). This is good for future per-corpus configuration, as the others can be computed very quickly.

\ignore{
\begin{figure}[t]\centering\vspace{1mm}%
\begin{minipage}[b]{.47\linewidth}%
\includegraphics[width=\linewidth,trim={0 0 0 0},clip]{feature_importance.png}
\subcaption{top 24 features}\label{fig:featureimportance:24}
\end{minipage}\hspace*{3mm}
\begin{minipage}[b]{.47\linewidth}%
\includegraphics[width=\linewidth,trim={0 0 0 0},clip]
{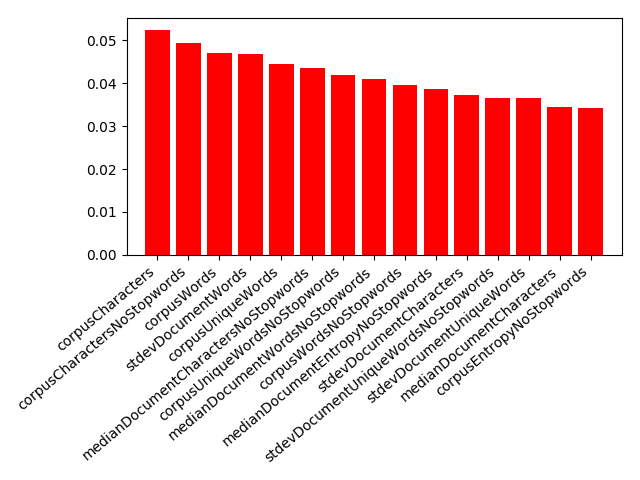} 
\subcaption{all 36 features}\label{fig:featureimportance:34}
\end{minipage}%
\vspace{-2mm}\caption{Gini importance, feature from Table~\ref{tab:corpusfeatures}. The subset on the left does not include language- and source-identifying features.}%
\label{fig:featureimportance}
\end{figure}
}


\begin{tcolorbox}
\textbf{Summary}: 
We can predict good configurations for unseen corpora reliably. 
Our predictions outperform the default configuration by 14\%, the best tuned single configuration by 4\%, and they are less than 1\% away from the virtual best solver.
\end{tcolorbox}


\section{Threats to Validity}
\label{sec:threats}

As with all empirical studies, there are a number of threats that may impair the validity of our results. 

Threats to construct validity concern the suitability of our evaluation metrics. Following many other works, we have used perplexity, the geometric mean of the inverse marginal probability of each word in a held-out set of documents~\cite{Hoffman2010}, to measure the fit of our topic models. Perplexity is not the only metric which can be used to evaluate topic models, and a study by Chang et al.~\cite{Chang2009} found that surprisingly, perplexity and human judgement are often not correlated. Future work will have to investigate the prediction of good configurations for textual software engineering corpora using other metrics, such as conciseness or coherence. The optimal may differ depending on the objective of the topic model, e.g., whether topics are shown to end users or whether they are used as input for another machine learning algorithm. In addition, selecting different corpus features might have led to different results. We selected easy-to-compute features as well as entropy as a starting point---studying the effect of other features is part of our future work.

Threats to external validity affect the generalisability of our findings. We cannot claim that our findings generalise beyond the particular corpora which we have considered in this work. In particular, our work may not generalise beyond GitHub README files and Stack Overflow threads, and also not beyond the particular programming languages we considered in this work. In addition, the amount of data we were able to consider in this work is necessarily limited. Choosing different documents might have resulted in different findings.

Threats to internal validity relate to errors in implementation and experiments. We have double-checked our implementation and experiments and fixed errors which we found. Still, there could be additional errors which we did not notice.

\section{Related Work}
\label{sec:related}

We summarise related work on the application of topic modelling to software artefacts, organised by the kind of data that topic modelling was applied to. We refer readers to Agrawal et al.~\cite{Agrawal2016topicModelling} for an overview of the extent to which parameter tuning has been employed by software engineering researchers when creating topic models. To the best of our knowledge, we are the first to explore whether good configurations for topic models can be predicted based on corpus features.

\subsection{Topic modelling of source code and its history}

In one of the first efforts to apply topic modelling to software data, Linstead et al.~\cite{Linstead2007} modelled Eclipse source code via author-topic models with the goal of mining developer competencies. They found that their topic models were useful for developer similarity analysis. Nguyen et al.~\cite{Nguyen2011} also applied topic modelling to source code, but for the purpose of defect prediction. The goal of their work was to measure concerns in source code, and then use these concerns as input for defect prediction. They concluded that their topic-based metrics had a high correlation with number of bugs.

With the goal of automatically mining and visualising API usage examples, Moritz et al.~\cite{Moritz2013} introduced an approach called ExPort. They found that ExPort could successfully recommend complex API usage examples based on the use of topic modelling. The goal of work by Wang and Liu~\cite{Wang2017} was to establish a project overview and to bring search capability to software engineers. This work also applied topic modelling to source code, and resulted in an approach which can support program comprehension for Java software engineers.

Thomas et al.~\cite{Thomas2014} focused their work on a subset of source code---test cases. The goal of their work was static test case prioritisation using topic models, and it resulted in a static black-box test case prioritisation technique which outperformed state-of-the-art techniques.

Applying topic modelling to source code history, Chen et al.~\cite{Chen2012}'s goal was to study the effect of conceptual concerns on code quality. They found that some topics were indeed more defect-prone than others. Hindle et al.~\cite{Hindle2011, Hindle2013} looked at commit-log messages, aiming to automatically label the topics identified by topic modelling. They presented an approach which could produce appropriate, context-sensitive labels to support cross-project analysis of software maintenance activities. Finally, Corley et al.~\cite{Corley2018} applied topic modelling to change sets with the goal of improving existing feature location approaches, and found that their work resulted in good performance.

\subsection{Topic modelling of bug reports and development issues}

Software engineering researchers have also applied topic modelling to bug reports and development issues, to answer a wide variety of research questions. In one of the first studies in this area, Linstead and Baldi~\cite{Linstead2009} found substantial promise in applying statistical text mining algorithms, such as topic modelling, for estimating bug report quality. To enable this kind of analysis, they defined an information-theoretic measure of the coherence of bug reports.

The goal of Nguyen et al.~\cite{Nguyen2012duplicate}'s application of topic modelling to bug reports was the detection of duplicates. They employed a combination of information retrieval and topic modelling, and found that their approach outperformed state-of-the-art approaches. In a similar research effort, Klein et al.~\cite{Klein2014}'s work also aimed at automated bug report deduplication, resulting in a significant improvement over previous work. As part of this work, the authors introduced a metric which measures the first shared topic between two topic-document distributions. Nguyen et al.~\cite{Nguyen2012inferring} applied topic modelling to a set of defect records from IBM, with the goal of inferring developer expertise through defect analysis. The authors found that defect resolution time is strongly influenced by the developer and his/her expertise in a defect's topic.

Not all reports entered in a bug tracking system are necessarily bugs. Pingclasai et al.~\cite{Pingclasai2013} developed an approach based on topic modelling which can distinguish bug reports from other requests. The authors found that their approach was able to achieve a good performance. Zibran~\cite{Zibran2016} also found topic modelling to be a promising approach for bug report classification. His work explored the automated classification of bug reports into a predefined set of categories.

Naguib et al.~\cite{Naguib2013} applied topic modelling to bug reports in order to automatically issue recommendations as to who a bug report should be assigned to. Their work was based on activity profiles and resulted in a good average hit ratio.

In an effort to automatically determine the emotional state of a project and thus improve emotional awareness in a software development team, Guzman and Bruegge~\cite{Guzman2013} applied topic modelling to textual content from mailing lists and Confluence artefacts. They found that their proposed emotion summaries had a high correlation with the emotional state of a project.

Layman et al.~\cite{Layman2016} applied topic modelling to NASA space system problem reports, with the goal of extracting trends in testing and operational failures. They were able to identify common issues during different phases of a project. They also reported that the process of selecting the topic modelling parameters lacks definitive guidance and that defining semantically-meaningful topic labels requires non-trivial effort and domain expertise.

Focusing on security issues posted in GitHub repositories, Zahedi et al.~\cite{Zahedi2018} applied topic modelling to identify and understand common security issues. They found that the majority of security issues reported in GitHub issues was related to identity management and cryptography.

\subsection{Topic modelling of Stack Overflow content}

Linares-V\'{a}squez et al.~\cite{Linares-Vasquez2013} conducted an exploratory analysis of mobile development issues, with the goal of extracting hot topics from Stack Overflow questions related to mobile development. They found that most questions included topics related to general concerns and compatibility issues. In a similar more recent effort, Rosen and Shihab~\cite{Rosen2016} set out to identify what mobile developers are asking about on Stack Overflow. They identified various frequently discussed topics, such as app distribution, mobile APIs, and data management. 

Looking beyond the scope of mobile development, Barua et al.~\cite{Barua2014} contributed an analysis of topics and trends on Stack Overflow. They found that topics of interest ranged widely from jobs to version control systems and C\# syntax. Zou et al.~\cite{Zou2015} applied topic modelling to Stack Overflow data with a similar goal, i.e., to understand developer needs. Among other findings, they reported that the most frequent topics were related to usability and reliability.

Allamanis and Sutton~\cite{Allamanis2013}'s goal was the identification of programming concepts which are most confusing, based on an analysis of Stack Overflow questions by topic, type, and code. Based on their work, they were able to associate programming concepts and identifiers with particular types of questions. Aiming at the identification of API usage obstacles, Wang and Godfrey~\cite{Wang2013} studied questions posted by iOS and Android developers on Stack Overflow. Their topic modelling analysis revealed several iOS and Android API classes which appeared to be particularly likely to challenge developers.

Campbell et al.~\cite{Campbell2013} applied topic modelling to content from Stack Overflow as well as project documentation, with the goal of identifying topics inadequately covered by project documentation. They were able to successfully detect such deficient documentation using topic analysis. As part of the development of a recommender system, Wang et al.~\cite{Wang2015} set out to recommend Stack Overflow posts to users which are likely to concern API design-related issues. Their topic modelling approach was able to achieve high accuracy.

\subsection{Topic modelling of other software artefacts}

Source code, bug reports, and Stack Overflow are not the only sources which researchers have applied topic modelling to. Other sources include usage logs, user feedback, service descriptions, and research papers. We briefly highlight related papers in this subsection.

Bajracharya and Lopes~\cite{Bajracharya2009, Bajracharya2012}'s goal was to understand what users search for. To achieve this, they mined search topics from the usage log of the code search engine Koders. They concluded that code search engines provide only a subset of the various information needs of users.

Aiming at the extraction of new or changed requirements for new versions of a software product, Galvis Carre\~{n}o and Winbladh~\cite{GalvisCarreno2013} applied topic modelling to user feedback captured in user comments. Their automatically extracted topics matched the ones that were manually extracted.

Nabli et al.~\cite{Nabli2018} applied topic modelling to cloud service descriptions with the goal of making it more efficient to discover relevant cloud services. They were able to improve the effectiveness of existing approaches.

In one of the first papers to report the application of topic modelling to software engineering data, Asuncion et al.~\cite{Asuncion2010} applied topic modelling to a variety of heterogeneous software artefacts, with the goal of improving traceability. They implemented several tools based on their work, and concluded that topic modelling indeed enhances software traceability.

Finally, Sharma et al.~\cite{Sharma2016} applied topic modelling to abstracts of research papers published in the Requirements Engineering (RE) conference series. Their work resulted in the identification of the structure and composition of requirements engineering research.

\section{Conclusions and Future Work}

Topic modelling is an automated technique to make sense of large amounts of textual data. To understand the impact of parameter tuning on the application of topic modelling to software development corpora, we employed techniques from Data-Driven Software Engineering~\cite{Nair2018DSE} to 40 corpora sampled from GitHub and 40 corpora sampled from Stack Overflow, each consisting of 1,000 documents. We found that (1) popular rules of thumb for topic modelling parameter configuration are not applicable to the corpora used in our experiments, (2) corpora sampled from GitHub and Stack Overflow have different characteristics and require different configurations to achieve good model fit, and (3) we can predict good configurations for unseen corpora reliably.

These findings play an important role in efficiently determining suitable configurations for topic modelling. State-of-the-art approaches determine the best configuration separately for each corpus, while our work shows that corpus features can be used for the prediction of good configurations. Our work demonstrates that source and context (e.g., programming language) matter in the textual data extracted from software repositories. Corpora related to the same programming language naturally form clusters, and even content from related programming languages (e.g., C and C++) are part of the same clusters. This finding opens up interesting avenues for future work: after excluding source code, why is the textual content that software developers write about the same programming language still more similar than textual content written about another programming language? In addition to investigating this, in our future work, we will expand our exploration of the relationship between features and good configurations for topic modelling, using larger and more diverse corpora as well as additional features and a longitudinal approach~\cite{McIntosh2018}. We will also make our approach available to end users through tool support and conduct qualitative research to determine to what extent the discovered topics make sense to humans.


\vspace{1mm}\noindent\textbf{Acknowledgments.} Our work was supported by the Australian Research Council projects DE180100153 and DE160100850. 
We acknowledge the support by the HPI Future SOC Lab, who granted us access to their computing resources.
\bibliographystyle{IEEEtran}
\bibliography{references}

\end{document}